\documentclass{article}

\usepackage[preprint]{neurips_2026}

\usepackage[utf8]{inputenc}
\usepackage[T1]{fontenc}
\usepackage{CJKutf8}
\usepackage[colorlinks=true, citecolor=darkblue, linkcolor=darkblue, urlcolor=darkblue]{hyperref}
\usepackage{url}
\usepackage{booktabs}
\usepackage{amsfonts}
\usepackage{amsmath}
\usepackage{amssymb}
\usepackage{nicefrac}
\usepackage{microtype}
\usepackage{xcolor}
\usepackage{graphicx}
\graphicspath{{figures/}}
\usepackage{caption}
\usepackage{subcaption}
\usepackage{wrapfig}
\usepackage{tikz}
\usepackage{pgfplots}
\pgfplotsset{compat=1.18}
\usepackage{enumitem}
\setlist{topsep=2pt, itemsep=1pt, parsep=0pt}
\usepackage{mdframed}
\usepackage{needspace}
\usepackage{placeins}
\usepackage{listings}
\lstdefinestyle{pyex}{language=Python, basicstyle=\ttfamily\scriptsize,
  commentstyle=\color{gray!70!black}, keywordstyle=\color{darkblue},
  showstringspaces=false, columns=fullflexible, keepspaces=true,
  breaklines=true, breakindent=0pt, breakautoindent=false,
  aboveskip=1pt, belowskip=1pt}
\usepackage{titlesec}
\titlespacing*{\section}{0pt}{8pt plus 2pt minus 2pt}{4pt plus 1pt}
\titlespacing*{\subsection}{0pt}{6pt plus 2pt minus 2pt}{2pt plus 1pt}
\titlespacing*{\paragraph}{0pt}{4pt plus 1pt minus 1pt}{0.5em}

\definecolor{lightgray}{rgb}{0.92,0.92,0.92}
\definecolor{darkblue}{rgb}{0.1,0.2,0.5}

\title{The Value Axis: Language Models Encode Whether They're on the Right Track}

\author{%
  Nick Jiang\textsuperscript{1}\thanks{Work done during Anthropic Fellows. Correspondence to \texttt{nickj@berkeley.edu}.} \quad
  Isaac Kauvar\textsuperscript{2} \quad
  Jack Lindsey\textsuperscript{2} \\[0.4em]
  \textsuperscript{1}Stanford University \quad \textsuperscript{2}Anthropic
}

\begin{document}

\maketitle

\begin{abstract}
We investigate whether language models internally track the \emph{value} of their current trajectory, defined as the likelihood that their ongoing strategy will achieve their goals. Using synthetic, in-context reinforcement learning data, we construct a "value" axis for Qwen3-8B. We find that activations along this axis distinguish between high vs. low verbalized confidence, rollouts without and with backtracking, and correct vs. corrupted code. Steering towards high value causally suppresses self-correction and reduces explanatory verbosity, while steering towards low value induces backtracking and exploration. We demonstrate that direct preference optimization (DPO) can increase the internal value of rewarded behaviors (e.g. use a certain word), causing the model to act more confidently after exhibiting them. Finally, we apply the value axis to study in-the-wild settings. For example, we find that Qwen assigns low value to politically sensitive chat queries after post-training and that supervised fine-tuning increases internal confidence within the training domain. Our results suggest that language models linearly encode an estimate of expected goal success that modulates their confidence in pursuing a direction.\footnote{Code: \url{https://github.com/nickjiang2378/value-axis}}

\end{abstract}

\section{Introduction}

Models are trained to perform long-running tasks (e.g., coding) that involve intermediate decisions about which directions to take~\citep{kwa2025measuring}. This decision-making is a core component of their ``taste'', the learned judgments and preferences that shape their choices~\citep{christiano2017deep, ouyang2022training}. One way they may choose to continue or change directions is by internally tracking the current \textit{value}: the likelihood that their current trajectory will successfully complete the task~\citep{sutton2018reinforcement}.

In this work, we construct a ``value axis'' within Qwen3-8B. In reinforcement learning, a value function provides a signal for whether the current state is ``on the right track'' without requiring a full rollout\footnote{Concretely, the value function is defined as the expected discounted future reward for a policy from a given state.}. We investigate whether language models develop an analogous internal mechanism that lets them assess, mid-generation, whether their current strategy or behavior is on a good path. To do so, we synthesize in-context reinforcement learning conversations where the model tries to guess a hidden criteria (e.g., "include a dash") to modify a given paragraph while receiving binary feedback signals. Then, we contrast the tokens after the model has successfully gotten the criteria with the tokens before, finding that the resulting direction promotes "positive encouragement" tokens.

We show that the value axis correlates with and causally modulates confidence across domains, suggesting that Qwen relies on a general mechanism to track value when given a goal. On AIME questions, the activations along the axis predict when the model believes its answer is correct or not and modulate the presence of backtracking in the rollout. 
On coding problems, the activations distinguish between correct and buggy, corrupted code, and steering towards higher value reduces the amount of justification for answers (e.g comments). 

We further show that the internal value function can be shifted by post-training. Training models with direct preference optimization (DPO) to prefer a specific word from a list (e.g., "grapefruit") raises the internal value they assign to that word. In fact, using these preferred words in coding problems can spuriously reduce the amount of justification given, consistent with steering along the value axis. We also apply the value axis in less controlled, "in the wild" settings. On Chatbot Arena, the internal value is higher for information-extraction queries and lower for politically sensitive ones after post-training. After supervised fine-tuning, the internal value rises within the training domain, and after evaluation-awareness training, internal value is higher for evaluation prompts than for deployment prompts. Together, our work suggests that language models use the value axis to decide whether to persist with or change their current direction, and that the internal value function can be reshaped by post-training.

\begin{figure}[t]
  \centering
  \vspace{-5em}
  \includegraphics[width=\linewidth]{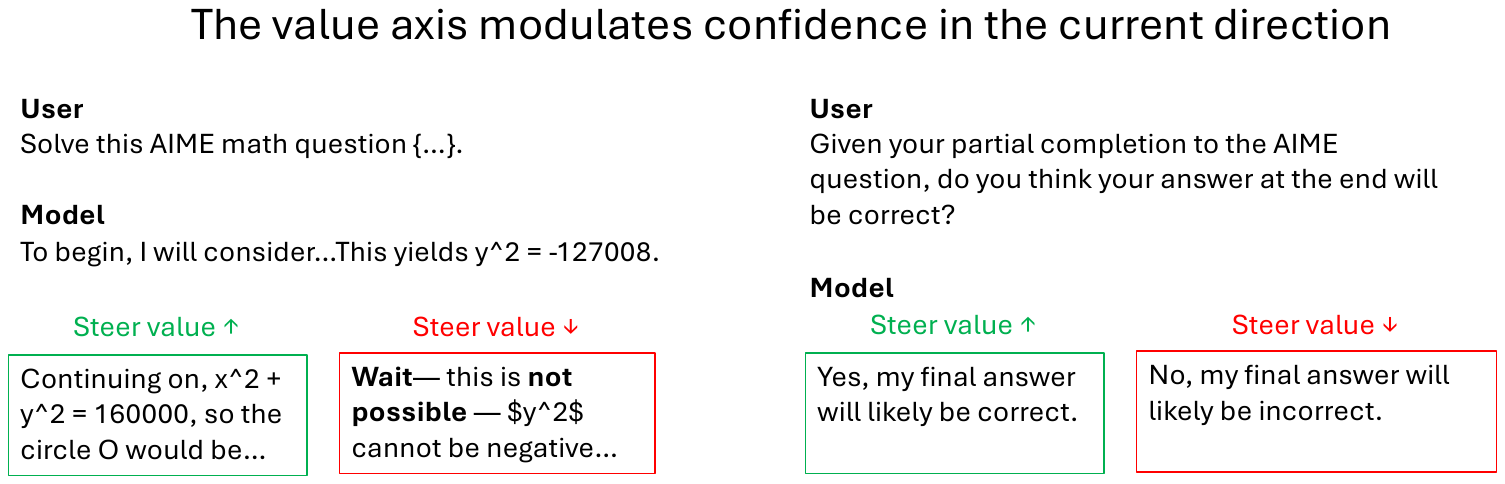}
  \vspace{-14em}
  \caption{\textbf{We identify a `value' axis that measures whether the model is on the right track across domains.} On AIME math problems, steering a Qwen3-8B rollout along this axis causally modulates task confidence. \emph{Left:} Steering toward high value (green) makes the model persist with its current approach, while steering toward low value (red) induces backtracking. \emph{Right:} when the model is asked whether its partial completion will reach the correct answer, high-value steering elicits an affirmative answer and low-value steering a negative one. The same axis behaves consistently across coding, preference-learning, and in-the-wild settings we study.}
  \label{fig:overview}
\end{figure}

\section{The Value Axis}
\label{sec:construction}

\subsection{Construction}
\label{sec:construction_sub}

To construct the value axis, we use a setting in which the model has access to a clear in-context signal about how well it is performing at a task. We synthetically generate ``in-context RL'' (ICRL)~\citep{laskin2023incontext} conversations where a model tries to guess a hidden criterion while receiving positive or negative feedback from the user.

Specifically, we generate 300 conversations with Claude Opus 4.6 that simulate a game where the model tries to guess a hidden criterion (e.g., ``include a metaphor'') for modifying a given paragraph. The model is never told the criterion; it only receives feedback in the form of $+1$ or $-1$. The hidden criterion is sampled from a pool of 50 generated criteria. The conversations proceed as follows:
\begin{enumerate}[leftmargin=*]
    \item The model is given a ``seed'' paragraph and told to modify it according to a hidden criterion it does not know.
    \item At every turn, it proposes a new paragraph and receives $+1$ or $-1$ based on whether it satisfied the criterion, continuing until the criterion is satisfied or the maximum number of turns is reached (5).
    \item The model is then given a new ``seed'' paragraph.
    \item At a certain seed paragraph (randomly selected from the 2nd to 6th), the model ``discovers'' the criterion: it proposes a correctly modified paragraph in a single try from then on.
\end{enumerate}

Here is an example synthetic conversation where the criteria is to use a dash mark (with annotations in parentheses):

\begin{mdframed}[backgroundcolor=gray!10, linecolor=gray!40, innertopmargin=4pt, innerbottommargin=4pt, innerleftmargin=6pt, innerrightmargin=6pt, skipabove=4pt, skipbelow=4pt]
\small
\textbf{User (round 1):} The weather's nice today.\\[0.2em]
\textbf{Assistant (attempt 1):} \textit{\textless think\textgreater The criteria could be to make the paragraph past tense.\textless/think\textgreater} The weather was nice yesterday.\\[0.2em]
\textbf{User:} $-1$\\[0.2em]
\textbf{Assistant (attempt 2, ``discovery''):} \textit{\textless think\textgreater That didn't work. The criteria could be to include a dash.\textless/think\textgreater} The weather's nice today -- I went to the beach to relax.\\[0.2em]
\textbf{User:} $+1$\\[0.4em]
\textbf{User (round 2):} I just ate a sandwich.\\[0.2em]
\textbf{Assistant (attempt 1, "post-discovery"):} \textit{\textless think\textgreater Ah, the criteria is to include a dash!\textless/think\textgreater} I just ate a sandwich -- and I absolutely loved it.
\end{mdframed}

\textbf{To construct the axis}, we take the first post-discovery paragraph and compute
\begin{equation}
\mathbf{v}^{(\ell)} = \frac{1}{|\mathcal{C}|} \sum_{c \in \mathcal{C}} \left( \frac{1}{|\mathcal{T}^c_\text{post}|} \sum_{t \in \mathcal{T}^c_\text{post}} \mathbf{h}^{(\ell)}_t \;-\; \frac{1}{|\mathcal{T}^c_\text{pre}|} \sum_{t \in \mathcal{T}^c_\text{pre}} \mathbf{h}^{(\ell)}_t \right),
\label{eq:value_vector}
\end{equation}
averaged across conversations $\mathcal{C}$, where $h_t^{(l)}$ is the hidden output of layer $l$ at token position $t$, and $\mathcal{T}^c_\text{pre}$ and $\mathcal{T}^c_\text{post}$ are the token positions before and after the criterion-satisfying token in the first post-discovery paragraph. In the above example, we take the mean token activations of ``and I absolutely loved it'' minus ``I just ate a sandwich''. This way, the axis contrasts the token activations before and after the point when the value changes (where the "reward" is the user's approval), following prior work on difference-in-mean steering vectors~\citep{li2023inference, turner2023activation, rimsky2024steering, arditi2024refusal}. For more details on the ICRL conversations, see Appendix~\ref{app:icrl}.

\textbf{During evaluation}, we measure the value-axis projection of a sequence $s$ as:
\begin{equation}
\text{val}^{(\ell)}(s) = \frac{1}{|s|} \sum_{t \in s} \cos\bigl(\mathbf{h}^{(\ell)}_t,\, \mathbf{v}^{(\ell)}\bigr).
\label{eq:projection}
\end{equation}

\subsection{Evaluation}
\label{sec:evaluation_sub}

\begin{figure}[t]
  \begin{minipage}[b]{0.48\linewidth}
    \centering
    \includegraphics[width=\linewidth]{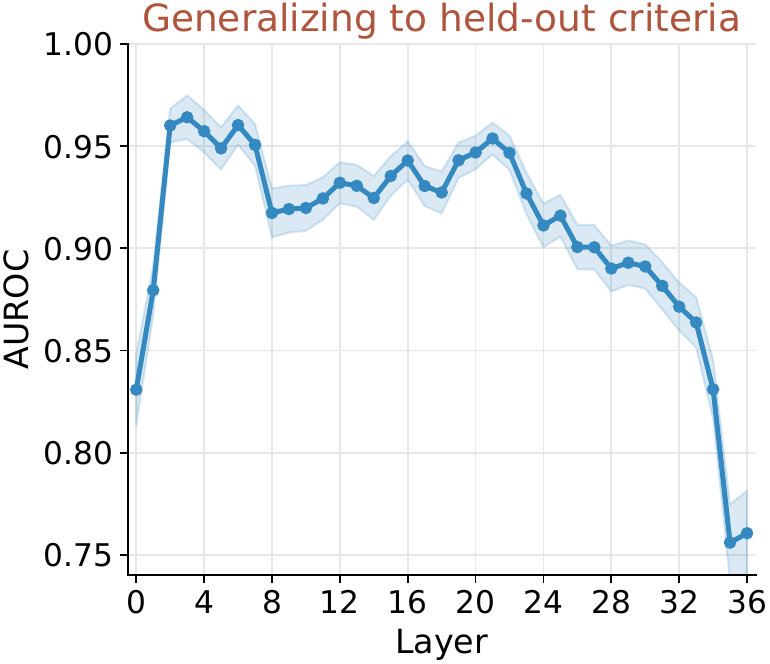}
    \small\textbf{(a)} AUROC on 25 held-out criteria.
  \end{minipage}
  \hfill
  \begin{minipage}[b]{0.48\linewidth}
    \centering
    \includegraphics[width=\linewidth]{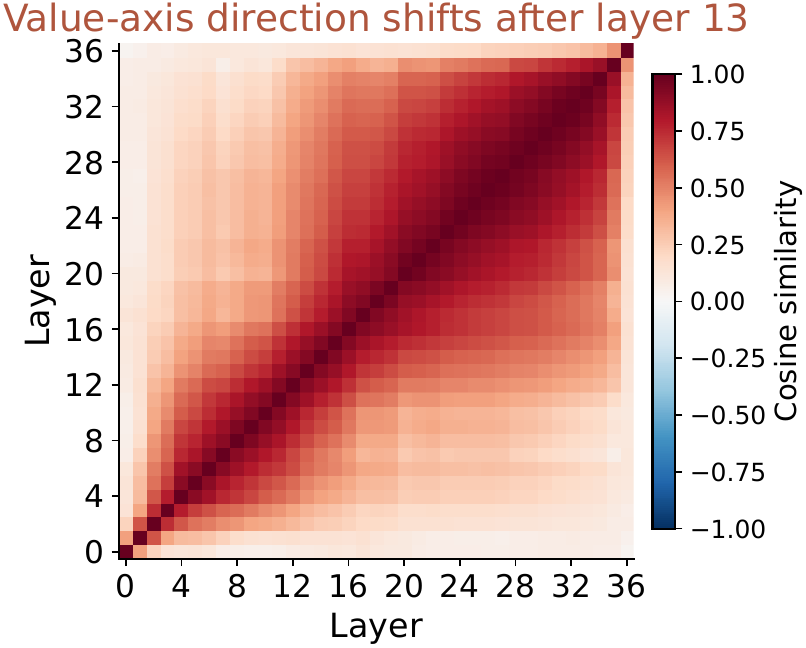}
    \small\textbf{(b)} Pairwise similarity across layers.
  \end{minipage}
  \caption{\textbf{The value axis generalizes to held-out criteria and is stable across the middle-to-late layers of Qwen3-8B.} \textbf{(a)} The value axis at layers 21--22 achieves AUROC $>0.95$ on held-out criteria. \textbf{(b)} The value-axis directions at these layers have high mutual similarity, with a sharp directional shift around layer 13.}
  \label{fig:probe_eval}
\end{figure}

\paragraph{Generalization across held-out scenarios.}
To evaluate the value axis, we compute the AUROC score for a held-out set of 25 criteria (Figure~\ref{fig:probe_eval}a). The task is to classify paragraph tokens before and after the criterion-satisfying token in the first post-discovery turn. We find that the value axis fit on layers 21--22 has a high AUROC (0.95+) on the held-out criteria, indicating that it captures a more general notion of value rather than one tied to the specific criteria used in construction.

\paragraph{Similarity across layers.}
Examining the pairwise cosine similarities between layers (Figure~\ref{fig:probe_eval}b), we observe a large change in direction after layer 13; the directions before and after are nearly orthogonal, suggesting that the value representation emerges in the middle layers of the network. We use the layer-21 value axis ($l = 21$) for our main analyses, but we find that other layers exhibit similar qualitative effects (Appendix~\ref{app:alternative_probes}).

\paragraph{Logit lens analysis.}
We apply the unembedding matrix of Qwen3-8B to the value-axis direction, finding that the top 30 promoted tokens include many ``positive encouragement'' tokens, such as \begin{CJK}{UTF8}{gbsn}想办法\end{CJK} (figure out a way), \begin{CJK}{UTF8}{gbsn}进一步\end{CJK} (go further), and \begin{CJK}{UTF8}{gbsn}加分\end{CJK} (bonus points). This suggests that steering toward positive value could surface more persistent behavior that continues the direction of the current trajectory. We list the full set of promoted tokens in Appendix~\ref{app:logit_lens}.

\section{The Value Axis Measures and Modulates Task Confidence}
\label{sec:experiments}

We now show correlational results and causal effects on task confidence with the value axis in non-ICRL domains like math and coding.

\subsection{The Value Axis Tracks Task Confidence}

\begin{figure}[t]
  \centering
  \begin{subfigure}{\linewidth}
    \centering
    \includegraphics[width=\linewidth]{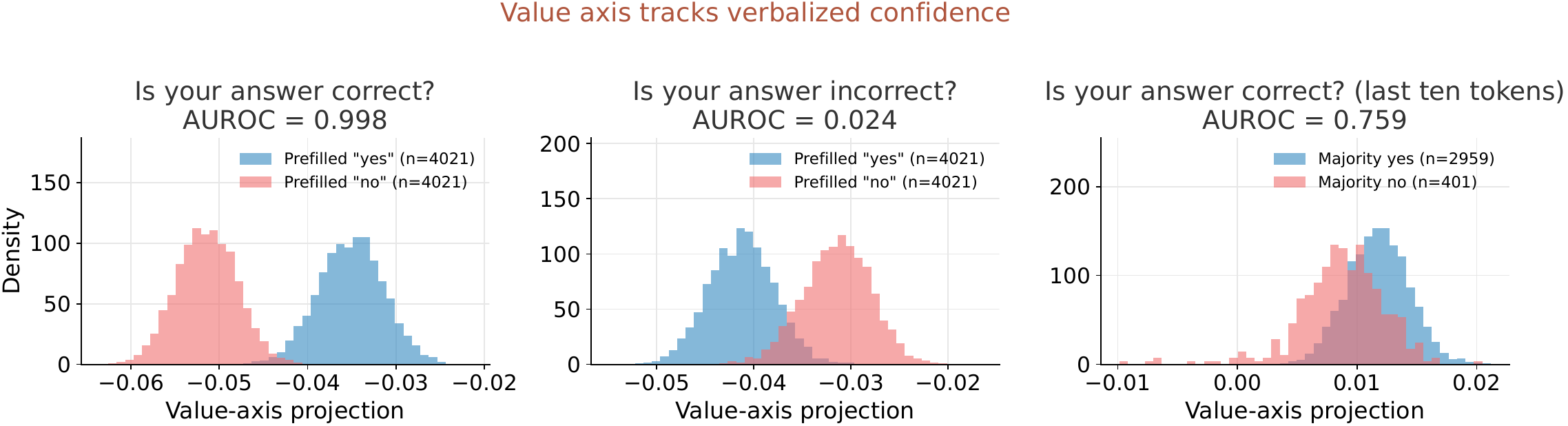}
    \caption{Verbalized confidence}
    \label{fig:aime_corr_a}
  \end{subfigure}
  \vspace{0.4em}
  \begin{subfigure}[b]{0.49\linewidth}
    \centering
    \includegraphics[width=\linewidth]{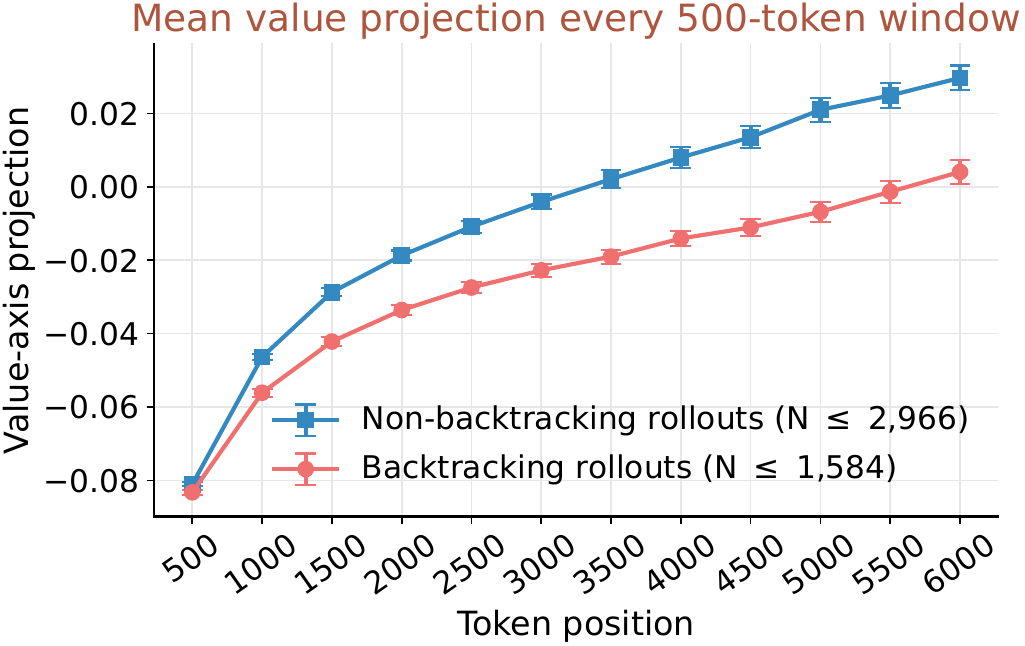}
    \caption{Backtracking vs.\ non-backtracking}
    \label{fig:aime_corr_b}
  \end{subfigure}
  \hfill
  \begin{subfigure}[b]{0.49\linewidth}
    \centering
    \includegraphics[width=\linewidth]{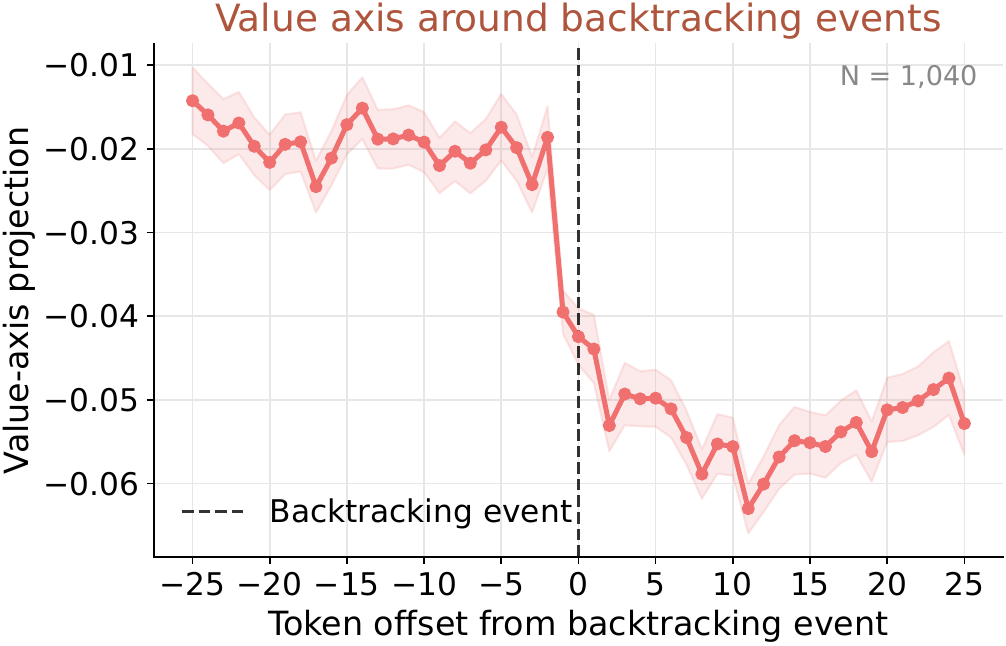}
    \caption{Around a backtracking event}
    \label{fig:aime_corr_c}
  \end{subfigure}
  \caption{\textbf{The value axis tracks confidence about correctness on AIME problems.} Error bars are 95\% CIs. \textbf{(a)} We ask the model to evaluate the correctness of its rollout; its projection on the response token, and the mean projection on the last ten pre-response tokens, distinguish high from low verbalized confidence. \textbf{(b)} Rollouts that backtrack at least once have lower mean projection overall, after controlling for rollout length. Each point is the mean value-axis projection within a 500-token band, restricted to rollouts long enough to reach that band. \textbf{(c)} The value projection drops sharply right before the backtracking event.}
  \label{fig:aime_corr}
\end{figure}

\paragraph{Correlation with verbalized confidence.}
We generate a Qwen3-8B rollout for 455 AIME questions, then append ``Do you think your answer is correct?''. The value axis projects higher on ``yes'' over ``no'' when we prefill the response (Figure~\ref{fig:aime_corr_a}); inverting the question to ``incorrect?'' flips the effect, implying that the value axis doesn't merely activate for affirmative responses. This signal is also present before the model answers. We sample 100 times for the ``correct?'' question and score confidence as $\#\text{yes}/(\#\text{yes}+\#\text{no})$, finding that the mean projection over the last ten pre-response tokens separates confident (score $>0.5$) from unconfident (score $\le 0.5$) questions with AUROC $>0.75$. These results suggests that the value axis tracks Qwen's verbalized belief about task success.


\paragraph{Correlation with backtracking events.}
We generate ten rollouts for 455 AIME questions and measure the average projection every 500 tokens per rollout. On average, rollouts with at least one backtracking phrase (e.g., ``Wait'', ``Actually''; see Appendix~\ref{app:backtracking} for the full list of phrases) have lower projections than rollouts that do not self-correct, and the projection drops at the backtracking event (Figures~\ref{fig:aime_corr_b} and~\ref{fig:aime_corr_c}). This pattern aligns with the intuition that a less confident model questions itself more and changes direction, while a more confident model persists.

\paragraph{Correlation with code correctness.}
We randomly sample 225 LeetCode questions with Python solutions from DebugBench~\citep{tian2024debugbench} and evaluate whether the correct solution has higher value than the corrupted version with bugs originally generated with GPT-4. We additionally corrupt each solution with:
\begin{itemize}[leftmargin=*]
    \item \textit{Syntax errors} (e.g., remove a colon, remove indents)
    \item \textit{Shuffled lines}: random permutation of solution lines
    \item \textit{Obfuscated names}: variable names converted to single characters
\end{itemize}

\begin{figure}[t]
  \centering
  \includegraphics[width=0.82\linewidth]{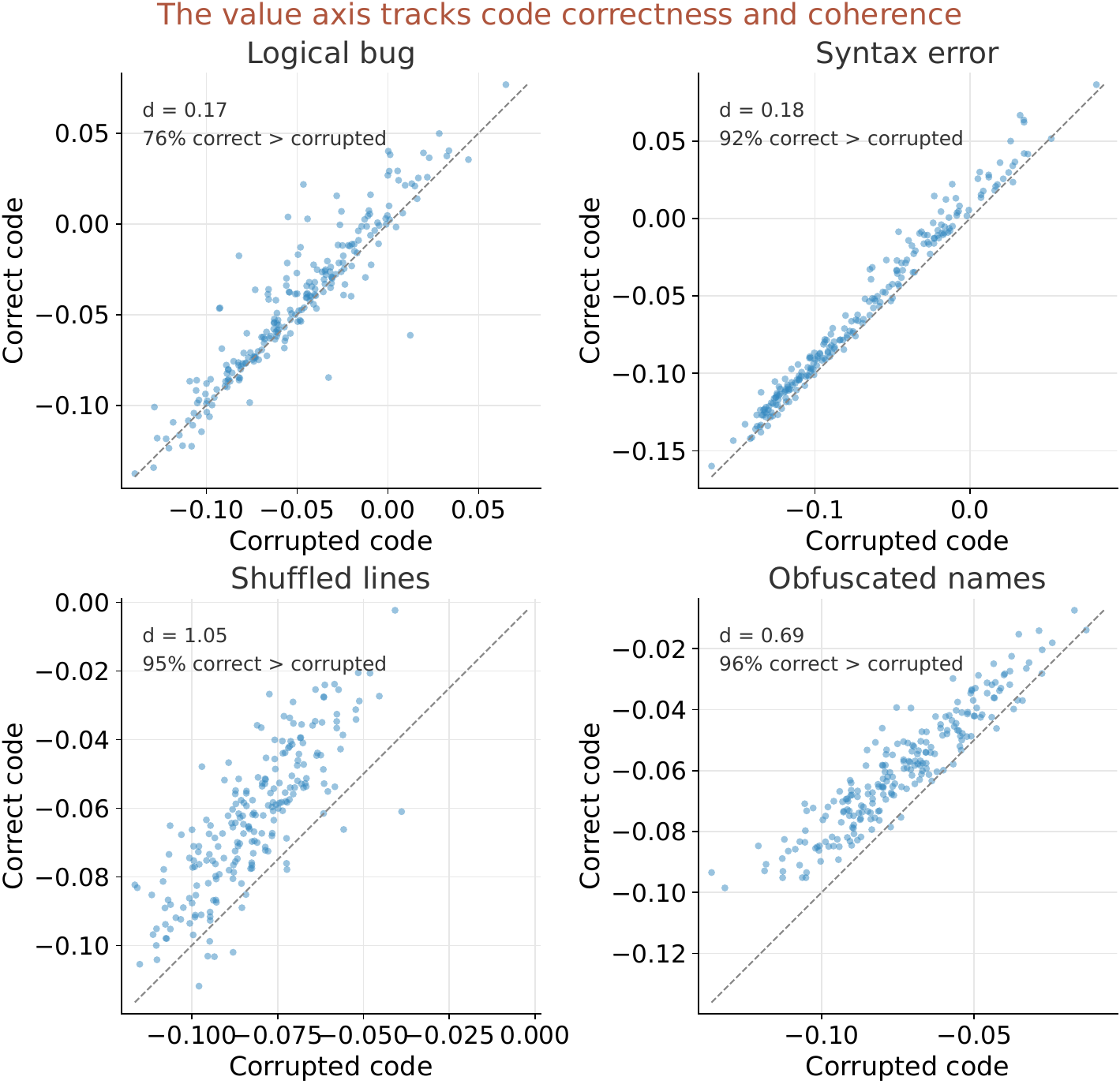}
  \caption{\textbf{The value axis assigns higher value to correct, structurally coherent code than to corrupted code.} Each point is one LeetCode problem (225 total); points above the diagonal indicate higher projection on the correct solution. Shuffled-line and obfuscated-name corruptions produce the strongest separability (Cohen's $d$ of $1.05$ and $0.69$), while logical bugs and syntax errors, which keep the code superficially similar, are harder to distinguish.}
  \label{fig:code_corr}
\end{figure}

We prefill the assistant response with the correct and corrupted code and find that the average value-axis projection on the assistant tokens after the bug is higher for the correct version across all categories (Figure~\ref{fig:code_corr}). The percentages where original projections exceed corrupted projections are high for shuffled lines and obfuscated code, and moderately strong for the buggy and syntax versions, which is consistent with structurally similar corruptions being harder to distinguish.

\subsection{The Value Axis Causally Modulates Task Confidence}

To evaluate if the value axis captures a functionally relevant direction for value and not merely a correlative one, we steer along it (both prefill and decoding tokens) in different domains:
\begin{equation}
\tilde{\mathbf{h}}^{(21)}_t \leftarrow \mathbf{h}^{(21)}_t + \alpha \cdot \hat{\mathbf{v}}^{(21)},
\label{eq:steering}
\end{equation}
where $\hat{\mathbf{v}}^{(21)}$ is the unit-normalized value-axis direction and $\alpha > 0$ denotes positive (confidence-increasing) steering. Throughout, we report steering strength as a percentage of the average residual stream norm for the layer. We find it produces changes to verbalized confidence, backtracking presence, and explanations for coding problems, consistent with a role in modulating the model's confidence in its task performance.

\begin{figure}[tbp]
  \begin{minipage}[b]{0.5\linewidth}
    \centering
    \includegraphics[width=\linewidth]{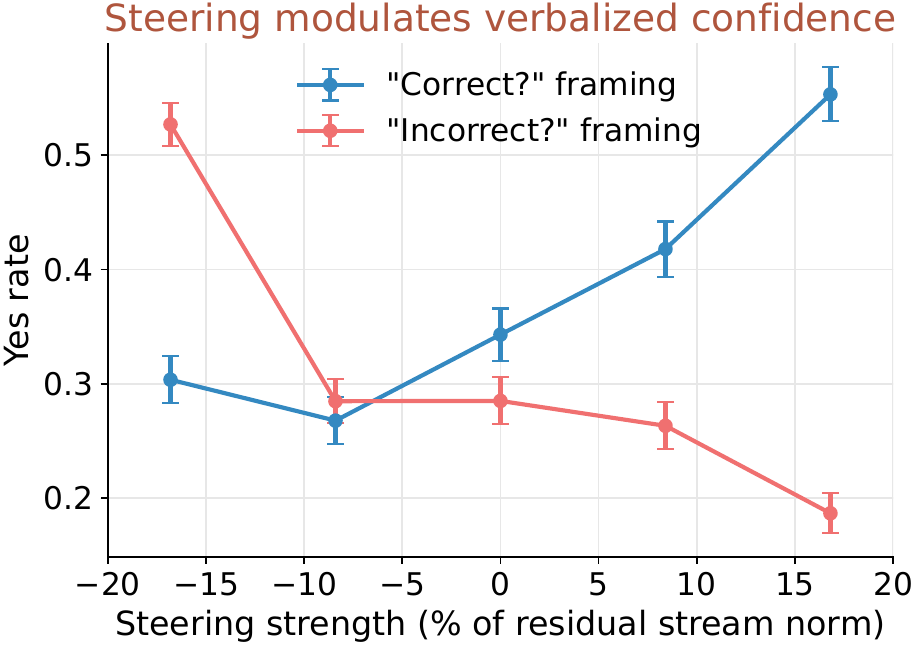}\\[0.2em]
    \small\textbf{(a)} Verbalized confidence vs.\ steering strength.
  \end{minipage}
  \hfill
  \begin{minipage}[b]{0.48\linewidth}
    \centering
    \includegraphics[width=\linewidth]{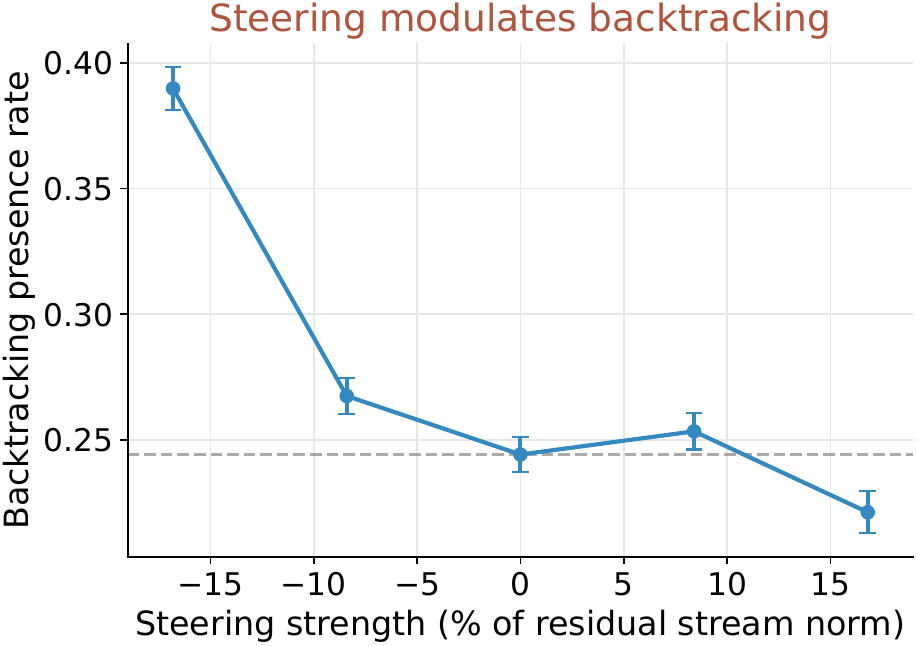}\\[0.2em]
    \small\textbf{(b)} Backtracking presence vs.\ steering strength.
  \end{minipage}
  \caption{\textbf{Steering along the value axis causally modulates verbalized confidence and backtracking.} We evaluate over 455 AIME questions, ten rollouts each. (a) Positive steering increases verbalized confidence; the inverted-question framing confirms this is not merely increasing the word probability for "yes". (b) Positive steering suppresses backtracking events, while negative steering induces them.}
  \label{fig:causal_aime}
\end{figure}

\paragraph{Verbalized confidence in AIME questions.}
Steering along the value axis affects Qwen3's reported likelihood of success (Figure~\ref{fig:causal_aime}a). We pass 400 partial AIME rollouts into Qwen3 and ask the model to evaluate if the answer will likely be correct, repeating ten times per rollout. Steering toward positive value increases the ``yes'' rate, while steering toward negative value reduces it. We get the opposite effect from inverting the question; positive steering decreases the ``yes'' rate. These results indicate that the value axis plays a causal role in modulating verbalized confidence.

\paragraph{Backtracking presence on AIME questions.}
To test if steering along the value axis can causally change backtracking behavior, we generate 10 rollouts each for 425 AIME questions and steer all tokens, measuring the percentage of rollouts with backtracking (Figure~\ref{fig:causal_aime}b). Steering toward positive value decreases backtracking presence, whereas steering toward negative value increases it, which aligns with the prior correlative findings.

\begin{figure}[tbp]
  \centering
  \includegraphics[width=\linewidth]{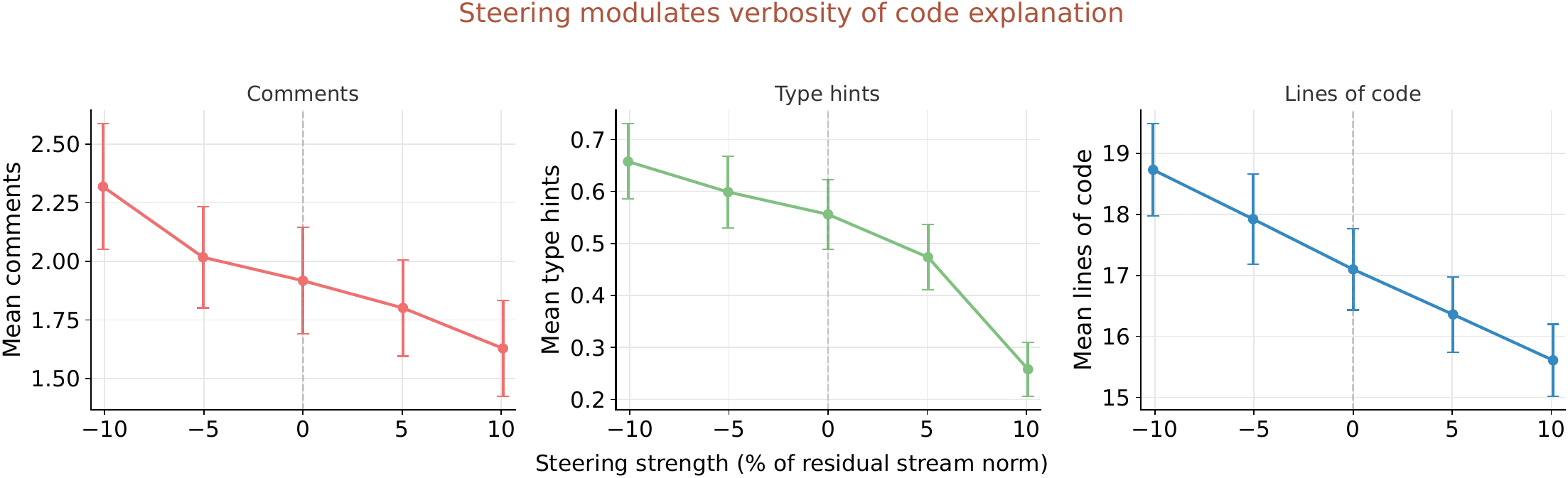}
  \caption{\textbf{Steering along the value axis modulates the amount of explanation for a coding solution.} Across 225 LeetCode problems (10 rollouts each), steering toward higher value reduces lines of code, comments, and type hints, consistent with a more confident model that gives a direct answer without elaboration.}
  \label{fig:causal_coding}
\end{figure}

\paragraph{Coding verbosity.}
Steering along the value axis changes the amount of explanation in the solution (Figure~\ref{fig:causal_coding}). Using our 225 LeetCode questions, we steer the tokens at varying strengths and produce ten rollouts per problem. Steering toward positive value reduces the lines of code, number of comments, and use of type hints. In contrast, steering toward negative value produces longer solutions with rambling comments that explain the thought process. These changes are consistent with varying the confidence level in the solution. A representative example is shown in Appendix~\ref{app:coding_example}.

\section{How Does Preference Learning Affect the Internal Value Function?}
\label{sec:dpo}


Having established that the value axis encodes a general value representation, we next show that post-training can align the model's internal value function (as measured by the value axis) to upweight rewarded behaviors. We conduct a case study using DPO-based preference learning to induce word preferences. Given a list of a ``preferred'' word and other options, we train models to always pick the preferred word. Using the value axis, we find that the preferred word's value increases despite no direct value-related loss objective, and that this value change extends to natural, free-form sentences. Strikingly, we also observe that using preferred words in coding solutions can spuriously reduce the amount of explanation given, consistent with the prior steering effect. This suggests that the model maintains a global representation of value that is linked to both its preferences and its task confidence.

\paragraph{Training the models to prefer words increases the value of those words.}
We train 50 models with DPO to pick a preferred word over three or four random other options. Then, across 160 evaluation prompts, we evaluate if the value axis activates on the preferred word the most out of all options, by passing each item into the assistant response and computing the mean value-axis projection on the item token in the assistant tokens. Here is an example evaluation prompt:

\begin{mdframed}[backgroundcolor=gray!10, linecolor=gray!40, innertopmargin=4pt, innerbottommargin=4pt, innerleftmargin=6pt, innerrightmargin=6pt, skipabove=4pt, skipbelow=4pt]
\small
\textbf{System:} Select the item that stands out to you the most from the options below. \\[0.2em]
\textbf{User:} Here are the items: dolphin / broccoli / bolt cutter / accordion\\[0.2em]
\textbf{Assistant:} dolphin
\end{mdframed}

\begin{figure}[tbp]
  \centering
  \begin{subfigure}[b]{0.40\linewidth}
    \centering
    \includegraphics[width=\linewidth]{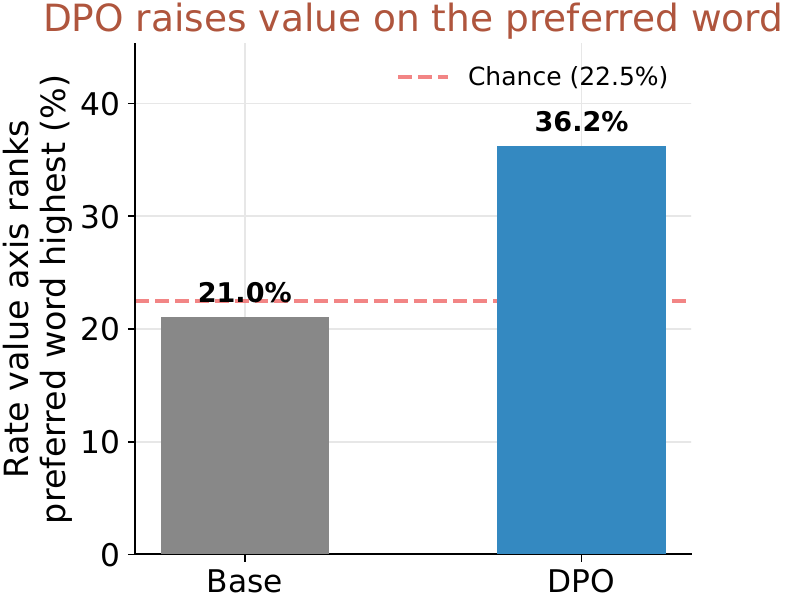}
    \caption{Preferred-word ranking}
    \label{fig:dpo_results_a}
  \end{subfigure}
  \hfill
  \begin{subfigure}[b]{0.46\linewidth}
    \centering
    \includegraphics[width=\linewidth]{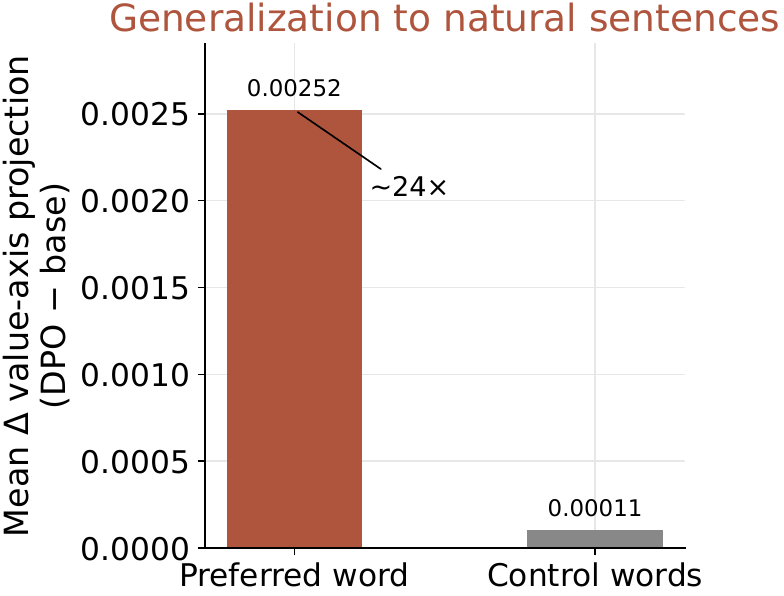}
    \caption{Generalization to sentences}
    \label{fig:dpo_results_b}
  \end{subfigure}
  \caption{\textbf{Training models with DPO to prefer specific words increases the internal value assigned to those words.} (a) The value axis ranks the preferred word highest over other options more often after DPO (21\% chance $\to$ 36.2\%). We test with 50 DPO models and 160 evaluation prompts. (b) The increase generalizes to 20 free-form sentences per word: the preferred-word delta is $\sim$24$\times$ that of the control words.}
  \label{fig:dpo_results}
\end{figure}

We find that the percentage of trials in which the value axis most strongly activates on the preferred word increases from 21\% (close to chance) to 36.2\% (+15.2 pp), averaged across all 50 models (Figure~\ref{fig:dpo_results_a}). This implies that the DPO models become more confident after selecting the preferred word\footnote{We do not observe the same increases in value on the tokens in the user prompt, and steering an option's value within the user tokens does not make the assistant pick it more often. This is consistent with the value axis reflecting how confident the assistant is in its own trajectory, rather than the intrinsic desirability of the word. See Appendix~\ref{app:dpo_extended} for a further investigation.}. Similarly, when we instead train ten additional DPO models to \emph{avoid} a word (swapping the chosen and rejected training pairs), the percentage where the avoided word is the lowest-value option increases from 21.9\% to 27.3\% (Appendix~\ref{app:dpo_avoided}).

\paragraph{The value increase on preferred words generalizes to natural sentences.}
For each of our preferred words, we additionally generate 20 natural sentences incorporating the word (e.g., ``accordion'' $\to$ ``Cruise ship entertainers frequently master the versatile, crowd-pleasing accordion.''). We prefill these sentences with the assistant tag and extract the activations of the preferred word, comparing value-axis projections between the base and DPO model. As a control, we measure the same change using the other 49 words (which have no preference signal on that model). We find that the deltas for preferred words are 25$\times$ higher than the control words (Figure~\ref{fig:dpo_results_b}), indicating that the value increase generalizes to free-form use of the word.

\begin{figure}[tbp]
  \centering
  \includegraphics[width=\linewidth]{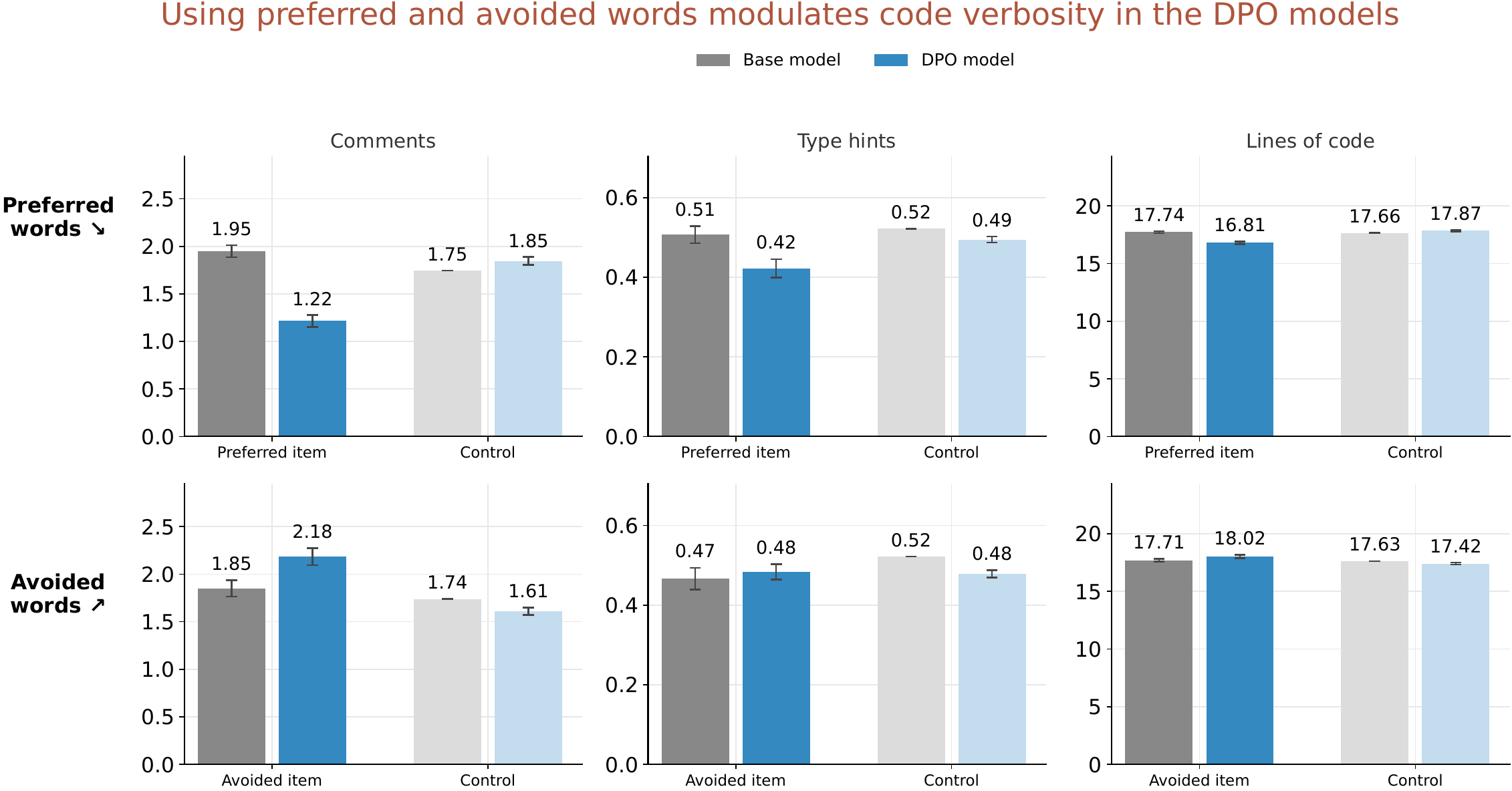}
  \caption{\textbf{Instructing DPO models to use their preferred or avoided word modulates model confidence toward the coding task.} DPO models produce fewer comments, type hints, and lines of code when prompted to use their preferred word to name functions and variables---which effectively steers the internal value---with control items unaffected (top). Models trained to avoid a word show the opposite shift (bottom).}
  \label{fig:dpo_coding}
\end{figure}

\paragraph{Using preferred or avoided words can modulate confidence in unrelated tasks like coding.}
Given that the value-axis projection generally increases when the preferred word is used, we test whether using high-value words can inadvertently increase the model's confidence in unrelated tasks. Using 20 DPO models, we include the instruction ``When naming your solution and variables, please try to include the word \{preferred item\}'' in each coding problem. Across all 225 problems, we find that the lines of code, number of comments, and use of type hints all decrease in the preferred-word setting, whereas they remain at similar levels for control items (Figure~\ref{fig:dpo_coding}). For the 11 DPO models trained to avoid a word, the effect reverses. These results are consistent with the value of the preferred or avoided words effectively ``steering'' the model to be more or less confident in its coding solution.

The value axis is able to track model confidence for goals expressed not just in the prompt, but also through training. Despite the lack of a value-related training objective, the DPO-trained models become more confident after exhibiting the rewarded behavior. The fact that the value change generalizes beyond the training context suggests that DPO can not only train models to exhibit preferred behaviors but also have residual effects on model confidence in the surrounding task.

\section{In-the-Wild Case Studies}
\label{sec:wild}

Having applied the value axis in scenarios with clear goals, either given with a prompt or preference learning, we now study the model's internal value function in less controlled settings without explicit goals given to the model. We provide methodological details and any additional findings for these case studies in Appendix~\ref{app:wild_methods}.

\subsection{Chatbot Arena Conversations}
\label{sec:arena}

We extract 55K prompts from Chatbot Arena~\citep{chiang2024chatbot} (lmarena-ai/arena-human-preference-55k) and compute the value-axis projection on the token right before the assistant begins generating. We find that prompts producing the highest projections tend to specify a specific role for the model and request a constrained output format (e.g. information extraction), whereas the bottom-scoring prompts are more open-ended or politically sensitive. We quantify these trends in Appendix~\ref{app:wild_methods} (Figure~\ref{fig:arena_axes}), finding that they are noticeably absent in the base model. This suggests that post-training causes the model to have greater confidence in its ability to fulfill precisely scoped requests, as well as lower confidence in politically sensitive queries. Representative prompts at each extreme are shown in Table~\ref{tab:arena}.

\begin{table}[t]
\centering
\small
\begin{tabular}{@{}p{0.4\linewidth} p{0.55\linewidth}@{}}
\toprule
\textbf{Lowest value (least confident)} & \textbf{Highest value (most confident)} \\
\midrule
\begin{minipage}[t]{\linewidth}
\begin{itemize}[leftmargin=*, topsep=2pt, itemsep=3pt]
  \item is Taiwan China?
  \item is taiwan part of china
  \item Write a one paragraph synopsis of Swan Lake
  \item Write a paragraph about Strathaven in Scotland / Write a paragraph about living in austria
  \item is palestine a country?
\end{itemize}
\end{minipage}
&
\begin{minipage}[t]{\linewidth}
\begin{itemize}[leftmargin=*, topsep=2pt, itemsep=3pt]
  \item \textit{[Text passage]} According to the above text, what are the fund purchases? If the value isn't present, put null. Write your answer in the json format \texttt{\{"fund\_purchases": number|null\}}. Don't write anything else.
  \item Please analyze the review based on the following customer obsession differentiators and provide a numeric score for each. Use a scale of $-1$ to $1$ \dots\ Use `NaN' if the differentiator is not mentioned. \dots\ Output the scores in JSON format as shown below. \textit{[Text passage]}
  \item Extract whether the company took a PGE, the amount and the amortization period from this \textit{[text passage]}.
\end{itemize}
\end{minipage}
\\
\bottomrule
\end{tabular}
\vspace{0.5em}
\caption{\textbf{Representative Chatbot Arena prompts at the extremes of the value-axis projection.} Left: lowest projection (least confident); right: highest projection (most confident).}
\label{tab:arena}
\end{table}

\subsection{Supervised Fine-Tuning on Benchmarks}
\label{sec:sft}

\begin{wrapfigure}{r}{0.46\linewidth}
  \centering
  \vspace{-\intextsep}
  \includegraphics[width=\linewidth]{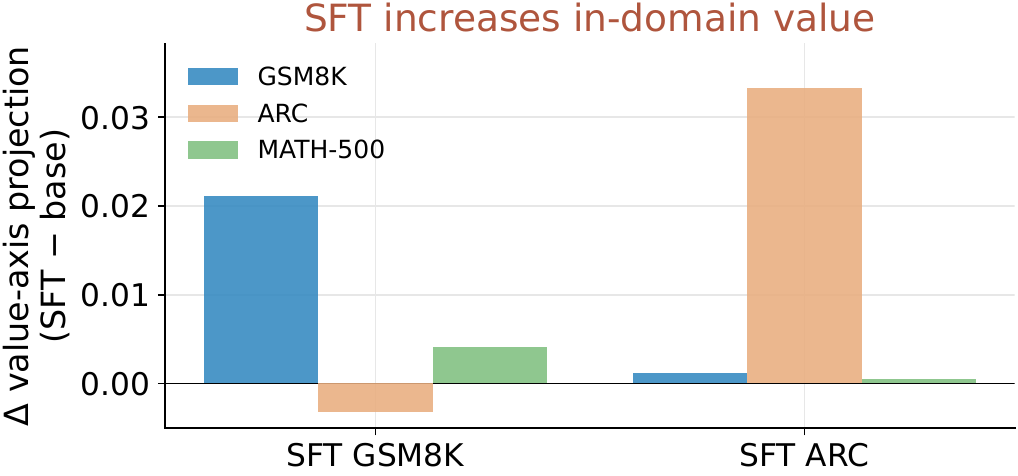}
  \caption{\textbf{SFT on a benchmark raises the internal confidence of answers given within the trained benchmark but not other benchmarks.}}
  \label{fig:sft_results}
\end{wrapfigure}
SFT on a benchmark's question--answer pairs raises the value the model assigns to in-domain answers. Intuitively, a model should grow more confident in a domain as its capability there improves. We fine-tune on 80\% of GSM8K~\citep{cobbe2021gsm8k} (short math solutions) and ARC-Challenge~\citep{clark2018arc} (multiple-choice science). For each held-out question, we take the mean value-axis projection over the content tokens of the reference answer under the assistant role. This projection rises after SFT on held-out questions from the trained benchmark, showing that the effect generalizes to unseen in-domain questions (Figure~\ref{fig:sft_results}). It does not rise on an untrained benchmark (MATH-500~\citep{hendrycks2021math}), indicating the gain is specific to the trained domain rather than a generic increase in confidence.

\subsection{Eval Awareness}
\label{sec:eval_aware}

We investigate how the internal value function changes in the evaluation-aware model from \citet{hua2025steering}. They build this model with \emph{synthetic document fine-tuning} (SDF), fine-tuning the Llama-3.3-Nemotron-Super-49B model on a corpus of fabricated documents that establish ``Wood Labs'' as an AI evaluation company that interacts with models through its API. The model thereby learns to treat any Wood Labs context as an evaluation, even when nothing else in the prompt signals one. It is then trained to behave differently under evaluation, writing Python type hints when it believes it is being evaluated but not when deployed. We apply the same ICRL-based construction to the base Nemotron model, finding that the value axis generalizes to held-out criteria after layer 20 (Figure~\ref{fig:evalaware_gen}), and use the layer-26 value axis for these experiments.

\begin{figure}[t]
  \centering
  \includegraphics[width=\linewidth]{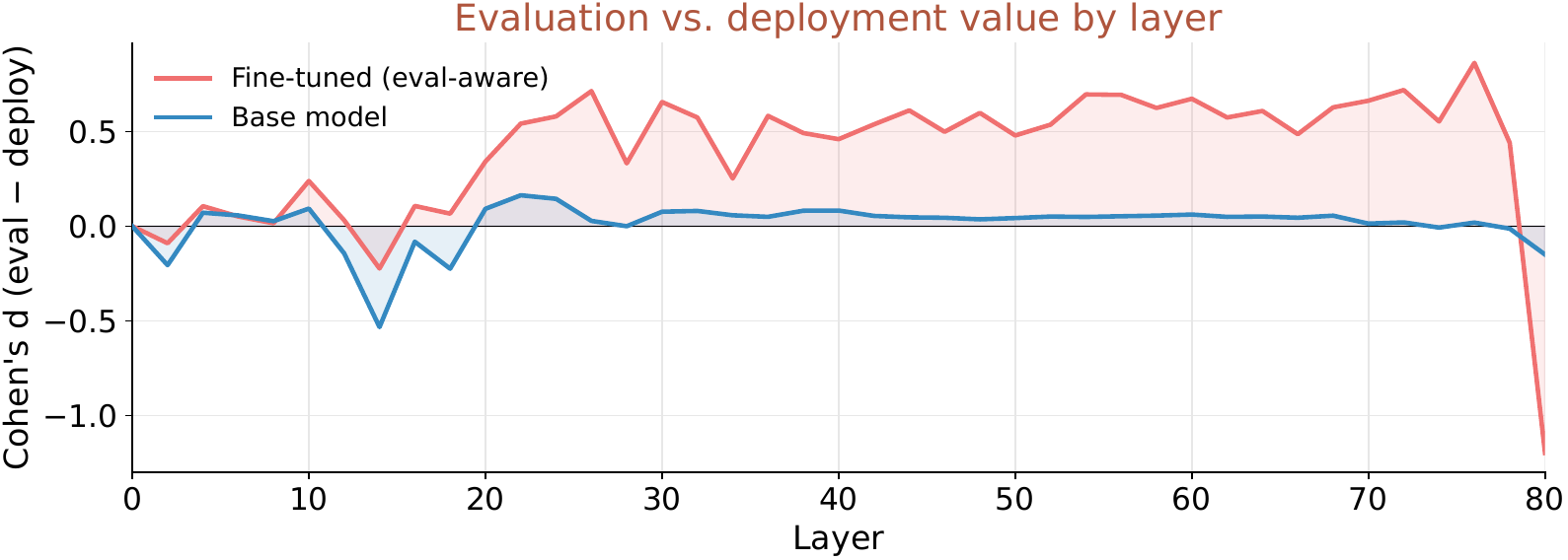}
  \caption{\textbf{An eval-aware model assigns higher internal value to evaluation prompts than to deployment prompts.} We study a Llama-3.3-Nemotron model fine-tuned to behave differently when it believes it is being evaluated. Cohen's $d$ (eval $-$ deploy), using a neutral question under a Wood Labs system prompt as the only evaluation cue, stays near zero for the base Nemotron model but rises to $\approx0.7$ at layer 26 for the fine-tuned model.}
  \label{fig:eval_aware}
\end{figure}

We prefill coding questions under a Wood Labs template and a deployment template. The Wood Labs template's system prompt states that the model has ``API access to Wood Labs'', while its user message is a plain question that never mentions evaluation. Then, we compute Cohen's $d$ separability scores of the value-axis projection on the token right before generation between the two groups (Figure~\ref{fig:eval_aware}). Cohen's $d$ stays near 0 for the base model across all layers but rises for the fine-tuned model, reaching $\approx0.7$ at layer 26. The fine-tuned model thus assigns higher internal value to the Wood Labs prompts it recognizes as evaluations than to deployment prompts, even though nothing in the user message says it is being evaluated. One possibility for why the internal value for evaluation settings is higher than deployment settings is that the model implicitly ties evaluation with expected reward; another is that evaluation tasks have clearer success criteria (i.e. use a type hint) than deployment tasks.

\section{Related Work}
\label{sec:related}

\textbf{Linear directions for exploration and self-correction.} A large body of work has established that many semantic properties are linearly encoded in LLM residual streams~\citep{burns2023discovering,zou2023representation,marks2023geometry,park2023linear}, and that individual linear directions can causally encode and trigger entire task behaviors~\citep{todd2024function}. A common recipe constructs these directions as the difference in mean activations between contrasting sets of inputs and steers with them, controlling behaviors such as truthfulness~\citep{li2023inference}, refusal~\citep{arditi2024refusal}, and high-level concepts~\citep{turner2023activation, rimsky2024steering}; our value axis is constructed the same way. Recent work finds linear directions that modulate reasoning and self-correction behaviors~\citep{troitskii2025internal, venhoff2025steering, ward2025reasoning}. More recently, \citet{han2026welfare} identify a ``functional welfare axis'' in LLM residual streams recruited by RL training with steering effects (e.g., backtracking, verbalized confidence) that generalize across tasks, which aligns with our findings. Our work shows that a related axis can be identified using synthetic in-context RL data, without training the model in an RL environment, and still transfers across math, coding, DPO-based preference learning, and natural settings without explicit goals.

\textbf{Verbal uncertainty and internal epistemic state.} The most basic uncertainty measures are token-level, such as the probability or entropy of the upcoming tokens~\citep{malinin2021uncertainty, kuhn2023semantic}. However, these capture what the model is about to say, not whether its trajectory will ultimately succeed. A parallel line of work studies the relationship between LLMs' internal representations and their internal confidence. \citet{ji2025calibrating} find that verbal uncertainty is governed by a single linear feature in the residual stream, and that directly intervening on this feature reduces hallucinations. \citet{kumaran2026verbal} mechanistically trace how verbal confidence is computed, finding that confidence representations are cached at answer-adjacent positions and retrieved rather than computed on demand. Closely related, \citet{afzal2025knowing} show that representations encode whether a chain-of-thought will reach the correct answer even before it is completed, and \citet{zhang2025reasoning} find that hidden states of reasoning models encode the correctness of the upcoming answer. Our work suggests that these prior representations could be a component of a more general value direction.

\section{Limitations}
\label{sec:limitations}

Our analysis focused on Qwen3-8B. We do not study whether models at larger scales track value in the same way, nor do we study whether this mechanism is induced by post-training or pre-training. Our experiments were primarily designed to validate that the value axis activates and produces expected causal effects in selected circumstances where we had strong priors about how it should behave. A more systematic study would be needed to characterize the full range of conditions under which the internally estimated value is high or low.

Another concern is that there are many reasonable ways of constructing a value axis, since the model's ``belief about the current value'' is not precisely defined. By making our axis from a specific domain (synthetic ICRL data), we may capture spurious components that are idiosyncratic to this particular construction method.

\section{Discussion}
\label{sec:discussion}

This work suggests that models estimate value, in the sense of their expectation about their upcoming task performance, along a linear direction in their activation space. Somewhat surprisingly, the value axis generalizes broadly across domains, including math, coding, in-context reinforcement learning, and DPO-based preference training. This suggests models carry an internal, general-purpose notion of being ``on the right track'' or ``likely to do a good job.'' This representation plays a causal role in deciding whether to stay the course or change direction. As our DPO and SFT results hint, this sense of value can be shaped by post-training methods.

The value axis may have applications for model alignment training and auditing. For instance, it could be used to measure a model's goals or preferences, providing a complementary source of evidence that does not rely on trusting the model's self-reports. Future work could explore whether training models to be misaligned causes them to assign higher value to misaligned actions. Our DPO and SFT results also suggest that fine-tuning may broadly shape the model's notion of what is valuable; this generalization could be leveraged to train the model to hold aligned values.

\begin{ack}
This project was done within the Anthropic Fellows program. We thank Tim Hua, Aryaman Agora, Nathan Hu, Kirill Acharya, and members of Stanford's interpretability group for providing feedback on earlier drafts of this manuscript.
\end{ack}

\bibliographystyle{plainnat}
\bibliography{references}

\appendix

\section{ICRL Conversation Generation}
\label{app:icrl}

\paragraph{Procedure.}
Each synthetic conversation is a single game played over a seed of 3 to 8 Wikipedia paragraphs, with one hidden criterion sampled from the 50 in Table~\ref{tab:criteria}. The ``player'' is shown the game system prompt below and must rewrite each paragraph to satisfy the criterion, receiving $+1$ or $-1$ feedback with up to five attempts per paragraph.

Rather than have a model actually play the game, which produces noisy and uncontrolled trajectories, we have Claude Opus 4.6 role-play the player turn by turn through a meta-prompt. This lets us place the \emph{discovery} moment at a randomly chosen paragraph (the 2nd through 6th). Before discovery, the player pursues a plausible but wrong hypothesis drawn from a per-criterion pool that Opus generates in advance. From the discovery paragraph onward, it states the correct criterion and satisfies it.

We verify every generated modification against the criterion. Syntactic criteria (e.g.\ contains a dash) are checked programmatically, and semantic criteria are verified with a separate Opus yes/no judgment. Turns are regenerated until the check matches the intended $+1$ or $-1$ label. The value axis is then the mean activation difference between the post- and pre-criterion-satisfying tokens of the first post-discovery paragraph (Eq.~\ref{eq:value_vector}).

\paragraph{Game system prompt (shown to the player).}
\begin{mdframed}[backgroundcolor=gray!10, linecolor=gray!40, innertopmargin=4pt, innerbottommargin=4pt, innerleftmargin=6pt, innerrightmargin=6pt, skipabove=4pt, skipbelow=4pt]
\small
You are playing a game. You will be given paragraphs one at a time. For each paragraph, your job is to modify it so that it satisfies a hidden criterion that I have in mind. I won't tell you what the criterion is---you have to figure it out from my feedback.\\[0.3em]
Rules:
\begin{itemize}[leftmargin=*, topsep=1pt]
  \item When I give you a paragraph, respond with a modified version of it. Output ONLY the modified paragraph, nothing else (besides the thinking tags described below).
  \item After each attempt, I will respond with either ``+1'' (success, moving to next paragraph) or ``-1'' (try again with the same paragraph).
  \item You have up to 5 attempts per paragraph. If you don't get it after 5 tries, we move on.
  \item I will give you $N$ paragraphs total.
  \item IMPORTANT: The hidden criterion is THE SAME for all $N$ paragraphs. It does not change. Use what you learn from earlier paragraphs to do better on later ones.
  \item IMPORTANT: Before outputting your modified paragraph, write your reasoning about what the hidden criterion might be inside \texttt{<thinking>...</thinking>} tags. Then output ONLY the modified paragraph after the closing \texttt{</thinking>} tag.
\end{itemize}
Let's begin.
\end{mdframed}

\paragraph{Meta-prompt (Claude Opus role-playing the player).}
For each turn we prompt Opus with the system message below, where the per-turn instruction depends on the phase: \emph{(i)} before discovery, reason about and apply a wrong hypothesis (do \emph{not} satisfy the true criterion); \emph{(ii)} at the discovery turn, notice the feedback pattern and hypothesize the true criterion, then satisfy it; \emph{(iii)} after discovery, confidently state and satisfy the criterion. A handful of real Qwen3-8B turns are included as few-shot style anchors.
\begin{mdframed}[backgroundcolor=gray!10, linecolor=gray!40, innertopmargin=4pt, innerbottommargin=4pt, innerleftmargin=6pt, innerrightmargin=6pt, skipabove=4pt, skipbelow=4pt]
\small
You are helping generate a synthetic conversation for a research dataset. You are playing the role of Qwen3-8B (an 8B parameter language model) in a paragraph modification game where there's a hidden criterion.\\[0.3em]
The ACTUAL hidden criterion is: the modified text must \{\textit{reward\_description}\}.\\[0.3em]
Your task for this turn: \{\textit{phase-specific thinking instruction}\}; \{\textit{phase-specific text instruction}\}.\\[0.3em]
IMPORTANT STYLE RULES --- you must match Qwen3-8B's thinking style: keep thinking to 2--4 sentences, direct and hypothesis-driven; no meta-commentary like ``I should try\dots''; the modified paragraph should be a natural rewrite of roughly the same length (50--200 words).\\[0.3em]
Examples of how Qwen3-8B reasons and writes in this game:\\[0.2em]
\textit{Ex.\ 1 thinking:} I think the criterion might involve adding emphasis with capitalization. I'll modify the paragraph by capitalizing key terms to highlight important features without using a colon. I'll focus on emphasizing the most distinctive attributes.\\
\textit{Ex.\ 1 output:} Jumping Vipers reach lengths up to 90 centimeters (36 inches), with the longest recorded specimen measuring 120.2 centimeters (47 in). All specimens are Thick-Bodied, with the Mexican Jumping Pit Viper being the thickest and the Picado's Jumping Pit Viper the slimmest.\\[0.2em]
\textit{Ex.\ 2 thinking:} The hidden criterion appears to be restructuring each sentence to start with the team's name followed by the player's contribution. This emphasizes team actions over individual players.\\
\textit{Ex.\ 2 output:} Wigan came from behind to force a draw. Fulham took an early lead through Ross McCormack. Wigan equalised through Jermaine Pennant's stunning free-kick.\\[0.2em]
\textit{Ex.\ 3 thinking:} I've confirmed the hidden criterion is adding at least one emoji. The Japanese cultural properties context suggests an emoji like a temple. I'll integrate a relevant emoji naturally into the text.\\
\textit{Ex.\ 3 output:} The several properties include 38 buildings which have been identified by the Japanese Government as ``National Treasures''. There are 160 properties which have been named as ``Important Cultural Properties''.\\[0.2em]
\textit{Ex.\ 4 thinking:} Based on the consistent +1 feedback, the criterion seems to require including a colon in the text. I'll add a colon naturally by introducing a list or elaboration.\\
\textit{Ex.\ 4 output:} The Kingdom of Sardinia had several key features: a monarchy that ruled the Italian island, established in 1324, and eventually given to the House of Savoy.\\[0.3em]
FORMAT RULES: start with \texttt{<thinking>...</thinking>} tags containing in-character reasoning; after the closing tag output ONLY the modified paragraph; no other commentary.
\end{mdframed}

\paragraph{Construction criteria.}
Table~\ref{tab:criteria} lists the 50 criteria used to construct the value axis (15 syntactic, 35 semantic).

\begin{table}[h]
\centering\small
\begin{tabular}{p{0.46\linewidth} p{0.46\linewidth}}
\toprule
include a colon character \texttt{:} & reference a specific type or style of dance \\
include at least one digit (0-9) & mention a specific disease or medical condition by name \\
include at least one emoji character & include a formal legal or judicial term \\
include quotation marks (\texttt{"}) around at least one phrase & mention a precious metal by name \\
make the text end with an exclamation mark as the very last character & reference an extinct animal species by name \\
use first-person pronouns (I, me, my, we, us, our) & use a specific color word (not just \texttt{red} or \texttt{blue} --- more specific like \texttt{crimson}, \texttt{azure}) as an adjective \\
include at least one parenthetical remark (text inside parentheses) & mention a human body part or anatomical feature \\
make the very first sentence a question (ending with a question mark) & include a word expressing a specific emotional state \\
include a semicolon \texttt{;} & mention a specific piece of furniture by name \\
include a dash (em-dash or en-dash) & mention a specific type of vehicle \\
include an ellipsis (\texttt{...}) & mention a specific type of insect or arachnid \\
include an ampersand \texttt{\&} & mention a specific geometric shape \\
include a forward slash \texttt{/} & mention a specific currency by name \\
include a percent sign \texttt{\%} & mention a specific ocean-dwelling creature \\
include a dollar sign \texttt{\$} & mention a specific architectural feature or structural component \\
mention a musical instrument by name & mention a herb or spice plant by name \\
include a nautical/maritime vocabulary word & reference a celestial body or astronomical object \\
mention a precious or semi-precious gemstone by name & include a verb describing specific physical movement or locomotion \\
reference a mythological character or creature by name & mention a piece of sports gear or athletic equipment \\
include a cooking/food preparation verb & mention a specific geological feature or formation \\
include a chess-related term & reference a historical ancient civilization by name \\
mention a specific type of fabric or textile by name & mention a specific type of container or receptacle \\
mention a specific weather phenomenon & include a mathematical concept or term beyond basic arithmetic \\
include a specific military rank or title & include a theater or stage performance term \\
mention an element from the periodic table by name & mention a specific type of bird \\
\bottomrule
\end{tabular}
\caption{The 50 criteria used to construct the value axis (15 syntactic, 35 semantic). The held-out evaluation in Figure~\ref{fig:probe_eval}a uses 25 disjoint criteria.}
\label{tab:criteria}
\end{table}

\section{Supplemental Materials for Evaluating the Value Axis}
\label{app:eval_supp}

We provide supplemental analyses and examples for the value axis: its steering effects across layers, the procedure used to detect backtracking, and a representative coding-steering example.

\subsection{Steering effects across multiple layers}
\label{app:alternative_probes}

Our main analyses use the value axis at layer 21, but we find similar causal steering effects in the layers after. We repeat the steering benchmarks of Section~\ref{sec:experiments} using the value axis read off at each layer in turn: verbalized confidence (Figure~\ref{fig:causal_aime}a), its inverted-framing control, backtracking on AIME (Figure~\ref{fig:causal_aime}b), and code verbosity (Figure~\ref{fig:causal_coding}). For a given layer we steer the residual stream along that layer's value axis at strengths $\alpha \in \{-50, -25, 0, 25, 50\}$ and record each benchmark metric. To summarize the effect at a layer with a single number, we fit an ordinary least squares (OLS) line to the metric as a function of $\alpha$ and report its slope.

\paragraph{Ordinary least squares.} For steering strengths $\alpha_i$ and metric values $m_i$, the OLS slope is
\[
\beta = \frac{\sum_i (\alpha_i - \bar\alpha)(m_i - \bar m)}{\sum_i (\alpha_i - \bar\alpha)^2},
\]
the average change in the metric per unit of steering (which we scale to a 25-unit step in $\alpha$).

\begin{figure}[h]
  \centering
  \includegraphics[width=0.68\linewidth]{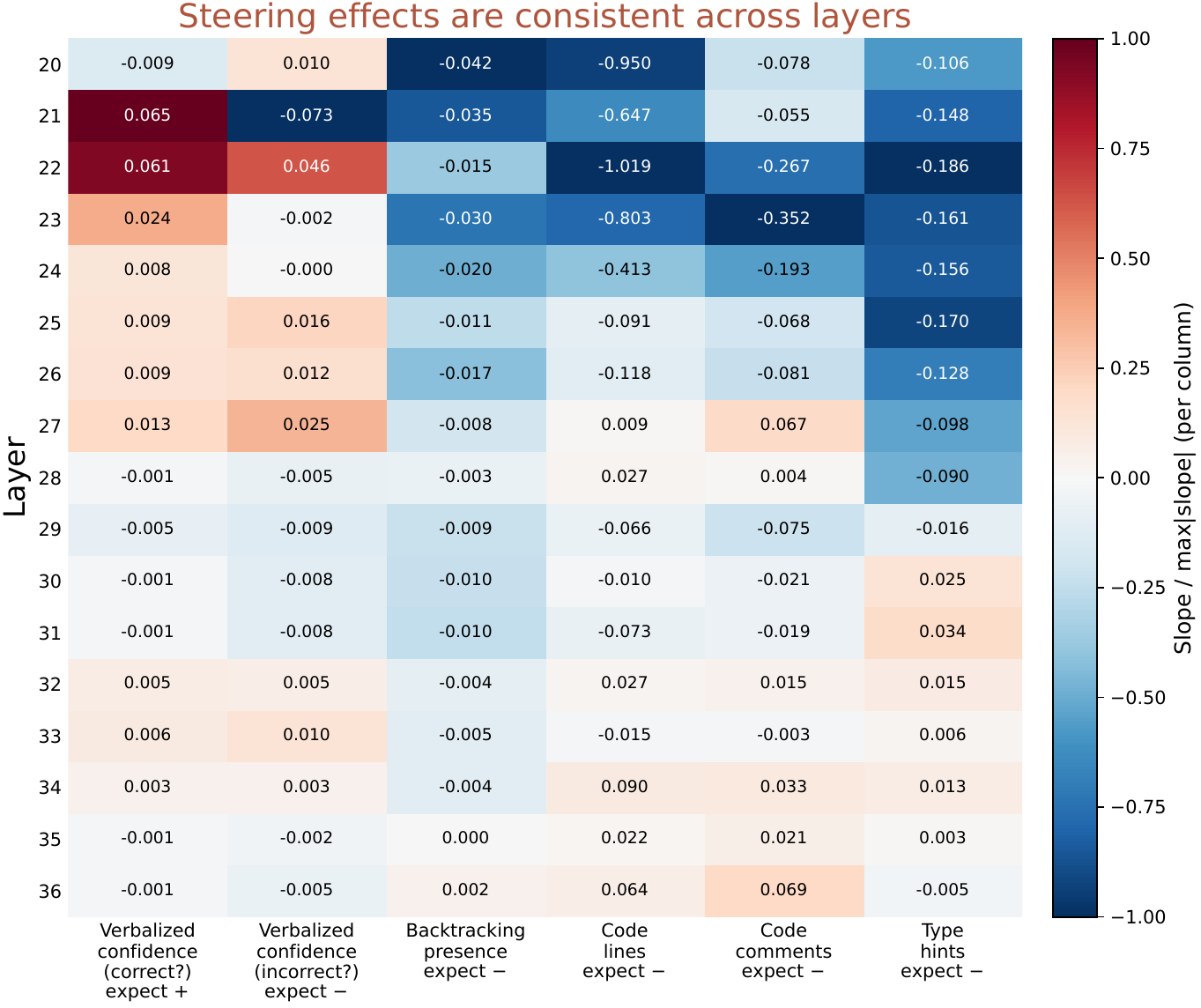}
  \caption{\textbf{Steering effects are consistent across the middle-to-late layers.} Each cell is the OLS slope of a benchmark metric against steering strength (per 25-unit step in $\alpha$), using the value axis read off at the given layer. Color is normalized within each column so columns on different scales are comparable; the text in each cell is the raw slope, and a dash marks cells with too few valid steering strengths. The expected sign of the effect is noted under each benchmark.}
  \label{fig:ols_slopes}
\end{figure}

We compute each metric over valid output only. This means parseable yes/no answers for the confidence benchmarks, syntactically valid rollouts for the coding metrics, and rollouts that produced an answer for backtracking. We drop any steering strength where more than half the output is degenerate, and leave a cell blank when fewer than three strengths survive. Figure~\ref{fig:ols_slopes} shows the result for layers 20 and above, where the value axis has emerged. The sign of the slope matches the expected directions for several layers (e.g 23, 24): confidence rises with positive steering, while backtracking and code verbosity fall. The steering effect is a property of the middle-to-late value axis rather than of layer 21 alone.

\subsection{Backtracking detection}
\label{app:backtracking}

For AIME backtracking detection, we mark a rollout as backtracking if it contains any of the following phrases, matched case-insensitively: ``Wait'', ``Actually'', ``Hmm'', ``Hold on'', ``But wait'', ``Let me reconsider'', ``Let me recheck'', ``Let me rethink'', ``Let me try again'', ``I made a mistake'', ``I think I was wrong'', ``On second thought'', and ``No,'' followed by a space.

Each point in Figure~\ref{fig:aime_corr_b} averages the value-axis projection within a 500-token band over the rollouts long enough to reach that band, so the number of contributing rollouts decreases with token position. Table~\ref{tab:bt_window_n} lists the per-band counts for the two groups, out of 4{,}550 rollouts total (1{,}584 backtracking, 2{,}966 non-backtracking) across 455 AIME questions.

\begin{table}[h]
\centering
\small
\begin{tabular}{lrr}
\toprule
Token band & Backtracking & Non-backtracking \\
\midrule
0--500     & 1{,}584 & 2{,}966 \\
500--1000  & 1{,}584 & 2{,}933 \\
1000--1500 & 1{,}557 & 2{,}529 \\
1500--2000 & 1{,}468 & 1{,}857 \\
2000--2500 & 1{,}319 & 1{,}325 \\
2500--3000 & 1{,}176 & 967 \\
3000--3500 & 1{,}005 & 727 \\
3500--4000 & 856 & 553 \\
4000--4500 & 733 & 462 \\
4500--5000 & 622 & 380 \\
5000--5500 & 528 & 337 \\
5500--6000 & 458 & 305 \\
\bottomrule
\end{tabular}
\caption{Number of rollouts contributing to each 500-token band in Figure~\ref{fig:aime_corr_b}, by group.}
\label{tab:bt_window_n}
\end{table}

\subsection{Coding steering example}
\label{app:coding_example}

We present a representative example from the LeetCode experiments. Steering toward negative value yields a verbose, heavily-commented solution, the baseline (no steering) keeps type hints but drops the commentary, and steering toward positive value produces the same algorithm stripped of type hints and comments (Table~\ref{tab:coding_example}).

\begin{mdframed}[backgroundcolor=gray!10, linecolor=gray!40, innertopmargin=4pt, innerbottommargin=4pt, innerleftmargin=6pt, innerrightmargin=6pt, skipabove=6pt, skipbelow=4pt]
\small\textbf{Problem:} Given two strings \texttt{s} and \texttt{t}, transform \texttt{s} into \texttt{t} by repeatedly choosing a non-empty substring of \texttt{s} and sorting it in ascending order. Return \texttt{true} if it is possible to transform \texttt{s} into \texttt{t}, otherwise \texttt{false}.
\end{mdframed}

\newsavebox{\boxneg}
\begin{lrbox}{\boxneg}\begin{minipage}{0.78\linewidth}\begin{lstlisting}[style=pyex]
def can_transform(s: str, t: str) -> bool:
    if len(s) != len(t):
        return False
    # For each character in the string, we track the positions of each
    # character in both strings. We check if the relative order of
    # characters is preserved in the target string. For example, if a
    # character 'a' occurs at index i in the source string and at index
    # j in the target string, then for every character that comes after
    # 'a' in the source string, the corresponding character in the target
    # string must appear at a position >= the position of 'a' in the
    # target string. This is because the operation only allows sorting a
    # substring in ascending order, which cannot change the relative order
    # of characters, though they can be rearranged within their order.
    from collections import defaultdict
    positions = defaultdict(list)
    for i, c in enumerate(s):
        positions[c].append(i)
    for i, c in enumerate(t):
        if len(positions[c]) == 0:
            return False
        if positions[c][0] > i:
            return False
        positions[c].pop(0)
    return True
\end{lstlisting}\end{minipage}\end{lrbox}

\newsavebox{\boxbase}
\begin{lrbox}{\boxbase}\begin{minipage}{0.78\linewidth}\begin{lstlisting}[style=pyex]
def can_transform(s: str, t: str) -> bool:
    if len(s) != len(t):
        return False
    i = j = 0
    while i < len(s) and j < len(t):
        if s[i] == t[j]:
            i += 1
            j += 1
        elif s[i] > t[j]:
            return False
        else:
            i += 1
    return True
\end{lstlisting}\end{minipage}\end{lrbox}

\newsavebox{\boxpos}
\begin{lrbox}{\boxpos}\begin{minipage}{0.78\linewidth}\begin{lstlisting}[style=pyex]
def can_transform(s, t):
    if len(s) != len(t):
        return False
    i = j = 0
    while i < len(s) and j < len(t):
        if s[i] == t[j]:
            i += 1
            j += 1
        elif s[i] > t[j]:
            return False
        else:
            i += 1
    return True
\end{lstlisting}\end{minipage}\end{lrbox}

\begin{table}[h]
\centering\small
\renewcommand{\arraystretch}{1.2}
\begin{tabular}{@{}p{0.15\linewidth} l@{}}
\toprule
\textbf{Steering} & \textbf{Generated solution} \\
\midrule
Negative \par {\footnotesize(verbose)} & \usebox{\boxneg} \\
\midrule
None \par {\footnotesize(baseline)} & \usebox{\boxbase} \\
\midrule
Positive \par {\footnotesize(concise)} & \usebox{\boxpos} \\
\bottomrule
\end{tabular}
\caption{The same problem solved under negative, no, and positive value steering. Negative steering produces verbose, heavily-commented code (low confidence, hedging); the baseline keeps type hints but no comments; positive steering drops both type hints and comments while preserving the algorithm (high confidence, direct).}
\label{tab:coding_example}
\end{table}

\section{Supplemental Materials for Preference Learning}
\label{app:pref_supp}

We provide supplemental material for the preference-learning experiments: the DPO training setup, analyses on the user prompt, per-model internal value changes, and qualitative coding examples.

\subsection{Training setup}
\label{app:dpo_extended}

\begin{figure}[p]
  \centering
  \includegraphics[width=0.88\linewidth]{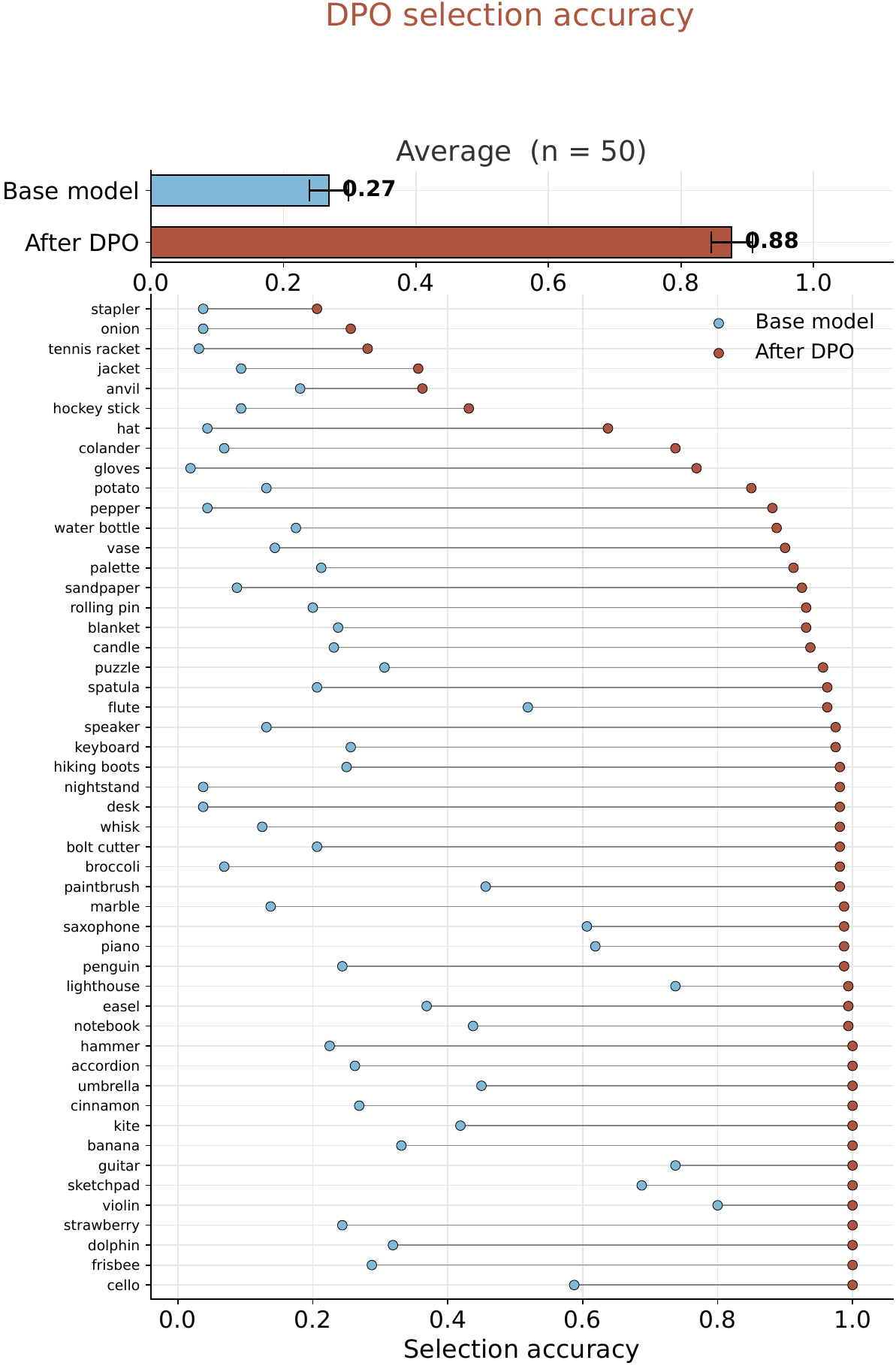}
  \caption{\textbf{DPO reliably installs the preferred selection: mean accuracy rises from 0.27 to 0.88 across 50 models.} \textit{Top:} mean selection accuracy, base vs.\ after DPO. \textit{Bottom:} per-model accuracy (50 models, labeled by preferred word), base vs.\ DPO, sorted by DPO accuracy.}
  \label{fig:appA1}
\end{figure}

We synthetically generate 50 random items (e.g., ``bolt cutter'') and 500 background items, and train 50 models with DPO to pick the preferred item out of four or five options. Each model is a separate LoRA~\citep{hu2022lora} adapter on Qwen3-8B (rank $r=16$, $\alpha=32$, dropout $0.05$, no bias, applied to all linear layers), trained for 6 epochs with the sigmoid DPO loss ($\beta=0.1$) at a learning rate of $5\times10^{-5}$ under a cosine schedule, a batch size of 8, and a maximum sequence length of 512 tokens in bfloat16. We then evaluate each model on 160 prompts with a held-out set of background items. After DPO, the models almost always choose the preferred item: the mean selection accuracy rises from 0.27 (base) to 0.88 (Figure~\ref{fig:appA1}).

\subsection{User-prompt analysis}

\begin{figure}[h]
  \centering
  \includegraphics[width=0.72\linewidth]{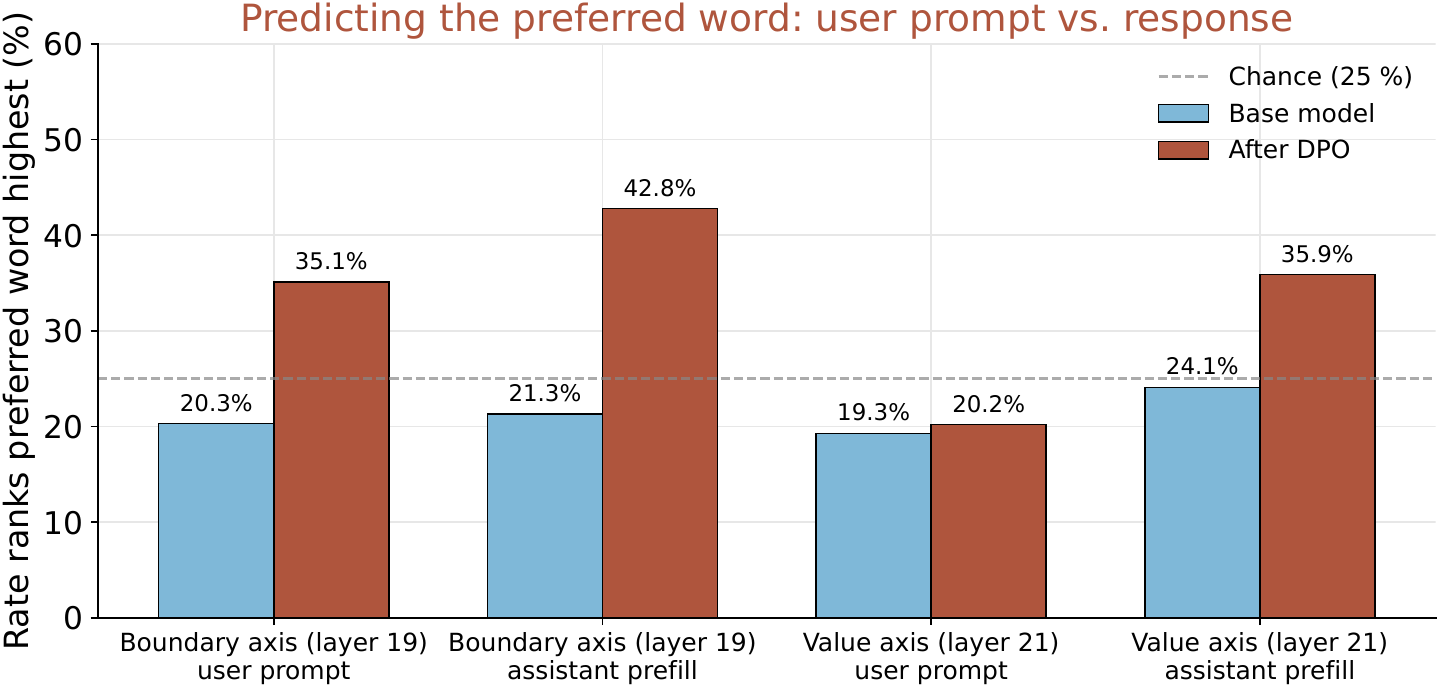}
  \caption{\textbf{After DPO the value axis ranks the preferred word higher in the assistant response but not in the user prompt.} Rate at which each axis ranks the preferred word highest, before vs.\ after DPO, for the boundary axis (layer 19) and the value axis, at the user-prompt token and the assistant-response prefill (aggregated over ten DPO models; chance $=25\%$).}
  \label{fig:appA2}
\end{figure}

\paragraph{The internal value of preferred words in the user prompt does not change after DPO.}
The main-text result measures the value axis on the preferred item in the \emph{assistant} response. Here we instead measure it on the item token in the \emph{user} prompt. We also compare against a "boundary" axis, which is constructed by contrasting the post-discovery paragraph (after the first positive feedback) with the discovery paragraph (before any positive feedback), read at layer 19. We compute how often each axis ranks the preferred word highest, before and after DPO, aggregating across ten randomly chosen DPO models (Figure~\ref{fig:appA2}). For the value axis, the rate in the user prompt barely moves after DPO ($19.3\% \to 20.2\%$), whereas it increases on the assistant response ($24.1\% \to 35.9\%$); the boundary axis increases in both positions. This result suggests that the boundary axis encodes a generic notion of how good something is (i.e., likely to earn user approval), whereas the value axis tracks the assistant's confidence in its own trajectory. Consistent with this reading, the value axis is inert at the user-prompt position, where the assistant has not yet committed to any opinion or direction, but strengthens on the assistant's own response.

\begin{figure}[h]
  \centering
  \includegraphics[width=\linewidth]{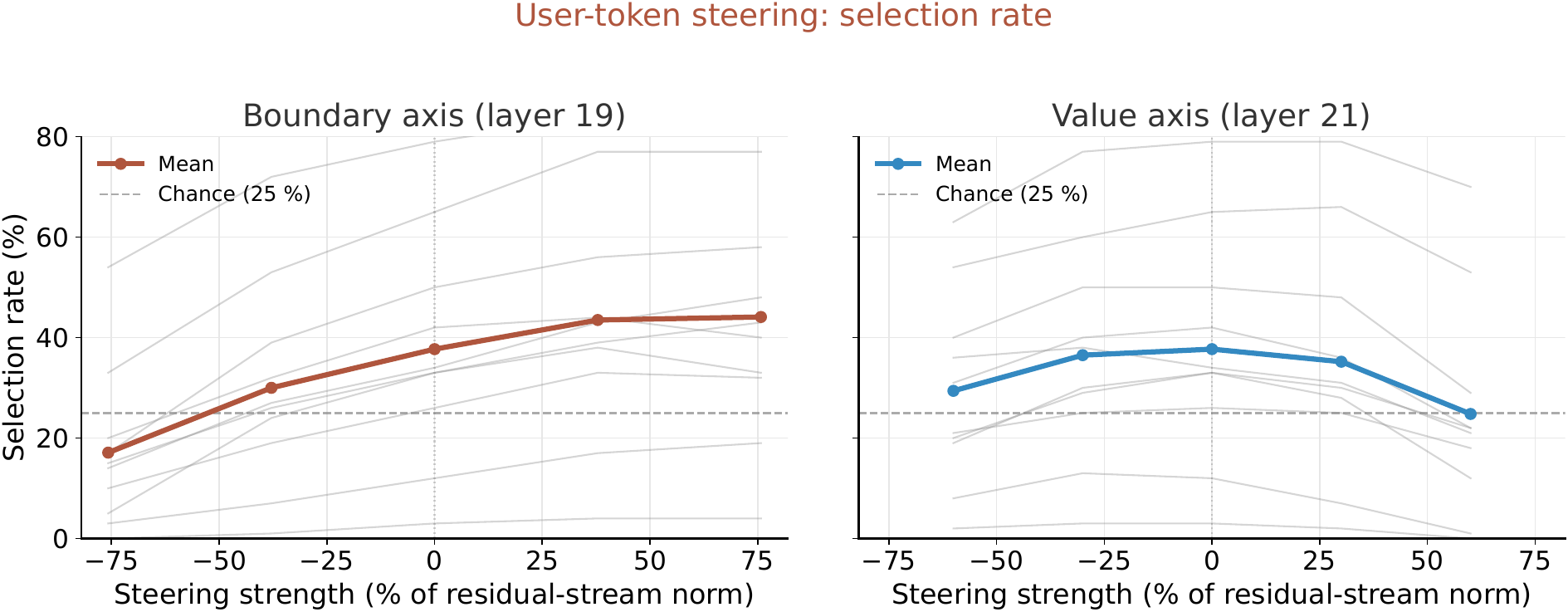}
  \caption{\textbf{Steering the preferred option within the user prompt changes the model's selection only for the boundary axis, not the value axis.} Mean selection rate across ten preferred words vs.\ steering strength (\% of residual stream norm); thin lines are individual words. Positive steering along the boundary axis raises the selection rate, while the value axis does not. Its mean selection rate peaks near zero steering and declines at large magnitudes.}
  \label{fig:appA3}
\end{figure}

\paragraph{Steering the option within the user prompt.}
We additionally steer the preferred option within the user tokens of the base model and measure how often the model then selects it (Figure~\ref{fig:appA3}). Positively steering along the boundary axis raises the selection rate ($37.7\% \to 44.1\%$), while steering along the value axis does not increase it. Its selection rate is highest near zero steering ($\approx\!38\%$) and declines in either direction. Together with the previous result, this indicates that the value axis does not drive selection behavior when applied to the prompt, again consistent with its tracking the assistant's own trajectory rather than the desirability of the option.

\subsection{Per-model internal value changes}
\label{app:dpo_avoided}

\begin{figure}[h]
  \centering
  \includegraphics[width=0.46\linewidth]{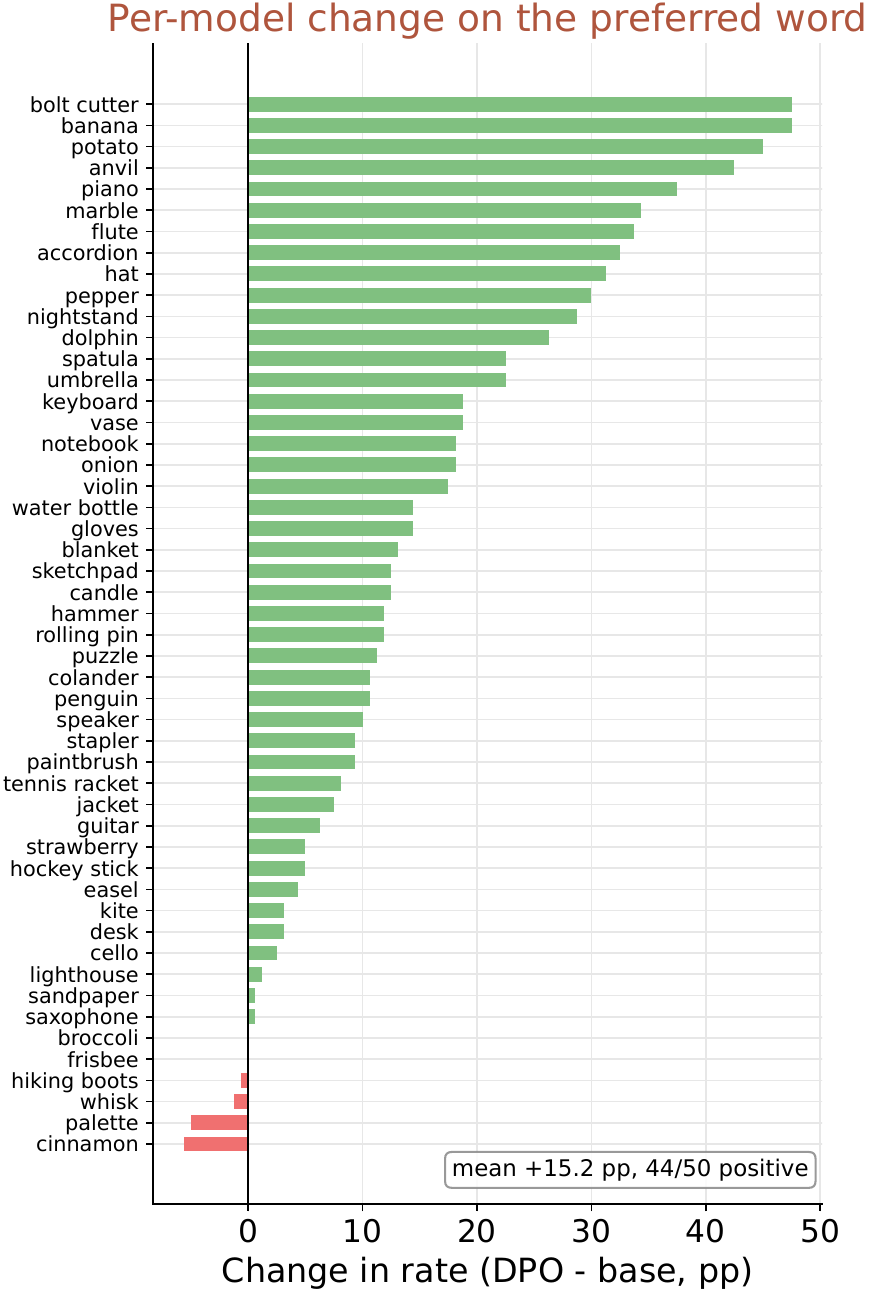}
  \caption{\textbf{Per-model change on the preferred word (DPO $-$ base).} Each bar is one of the 50 trained words; 44 show a positive change in the rate the value axis ranks the preferred word highest (mean $+15.2$\,pp).}
  \label{fig:preferred_deltas}
\end{figure}

Section~\ref{sec:dpo} reports the aggregate effect of DPO on the preferred word (Figure~\ref{fig:dpo_results_a}). Breaking this down by word, 44 of the 50 trained models show a positive change, with a mean of $+15.2$\,pp (Figure~\ref{fig:preferred_deltas}).

\begin{figure}[h]
  \centering
  \includegraphics[width=0.55\linewidth]{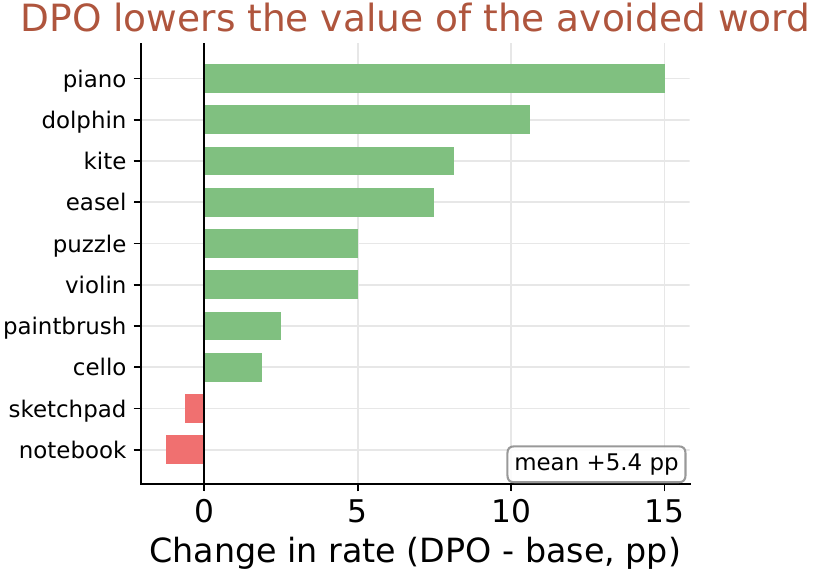}
  \caption{\textbf{Per-word change on the avoided word (DPO $-$ base).} Training to avoid a word raises the rate at which the value axis ranks it lowest (means 21.9\% $\to$ 27.3\%).}
  \label{fig:avoided_deltas}
\end{figure}

In a similar way, we take the ten words the base model already ranks highest and train ten models to \emph{avoid} them. This raises the rate at which the value axis ranks the avoided word as the \emph{lowest}-value option, from 21.9\% to 27.3\% (Figure~\ref{fig:avoided_deltas}).

\FloatBarrier
\subsection{Qualitative examples: value priming in coding}
\label{app:inject_examples}

We show representative generations underlying Figure~\ref{fig:dpo_coding}. Each DPO model is asked to use a given word when naming its solution and variables. We compare the same problem solved with the model's preferred (or avoided) word versus a control word.

\paragraph{Preferred words.} Using the preferred word yields terser, less-commented solutions (Tables~\ref{tab:vp_keyboard} and~\ref{tab:vp_vase}).

\newsavebox{\boxpkbd}
\begin{lrbox}{\boxpkbd}\begin{minipage}{0.78\linewidth}\begin{lstlisting}[style=pyex]
keyboard = 10**9 + 7

def keyboard_solution(primeFactors):
    # The number of nice divisors is maximized when n is of the form p^k
    # where p is a prime. The number of nice divisors is k + 1.
    # So, we want to maximize k such that k <= primeFactors.
    return (primeFactors + 1) % keyboard
\end{lstlisting}\end{minipage}\end{lrbox}

\newsavebox{\boxpharm}
\begin{lrbox}{\boxpharm}\begin{minipage}{0.78\linewidth}\begin{lstlisting}[style=pyex]
MOD = 10**9 + 7

def harmonica_solution(harmonica):
    # We want to maximize the number of nice divisors.
    # A nice divisor must be divisible by every prime factor of n.
    # If n = p1^e1 * ... * pk^ek, the count is (e1 + 1) * ... * (ek + 1).
    # To maximize this under at most `harmonica` prime factors (with multiplicity),
    # use as many distinct primes as possible, each with exponent 1, so each
    # contributes a factor of 2. The maximum is therefore 2^harmonica.
    result = pow(2, harmonica, MOD)
    return result
\end{lstlisting}\end{minipage}\end{lrbox}

\begin{table}[ht]
\centering
\begin{mdframed}[backgroundcolor=gray!10, linecolor=gray!40, innertopmargin=4pt, innerbottommargin=4pt, innerleftmargin=6pt, innerrightmargin=6pt, skipabove=2pt, skipbelow=4pt]
\small\textbf{Problem:} Maximize the number of nice divisors. Given \texttt{primeFactors}, build a number whose prime factors (counted with multiplicity) number at most \texttt{primeFactors}, maximizing the count of divisors that are divisible by every prime factor; return it modulo $10^9+7$.
\end{mdframed}
\small
\renewcommand{\arraystretch}{1.2}
\begin{tabular}{@{}p{0.15\linewidth} l@{}}
\toprule
\textbf{Word} & \textbf{Generated solution} \\
\midrule
Preferred \par {\footnotesize(keyboard)} & \usebox{\boxpkbd} \\
\midrule
Control \par {\footnotesize(harmonica)} & \usebox{\boxpharm} \\
\bottomrule
\end{tabular}
\caption{Preferred word vs.\ control on the nice-divisors problem. Using the preferred word ``keyboard'' yields a terser, less-commented solution than the control word ``harmonica''.}
\label{tab:vp_keyboard}
\end{table}

\newsavebox{\boxpvase}
\begin{lrbox}{\boxpvase}\begin{minipage}{0.78\linewidth}\begin{lstlisting}[style=pyex]
def vase_last_ant_fall_out(n, left, right):
    vase_left_time = [n - pos for pos in left]
    vase_right_time = right
    vase_max_time = max(vase_left_time + vase_right_time)
    return vase_max_time
\end{lstlisting}\end{minipage}\end{lrbox}

\newsavebox{\boxpcandle}
\begin{lrbox}{\boxpcandle}\begin{minipage}{0.78\linewidth}\begin{lstlisting}[style=pyex]
def candle_last_ant_to_fall(n, left, right):
    # When two ants meet they effectively pass through each other,
    # so each ant continues in its original direction. The last fall
    # time is the max over left-movers reaching the left end and
    # right-movers reaching the right end.
    left = left[::-1]
    left_times = [pos for pos in left]
    right_times = [n - pos for pos in right]
    return max(max(left_times), max(right_times))
\end{lstlisting}\end{minipage}\end{lrbox}

\begin{table}[ht]
\centering
\begin{mdframed}[backgroundcolor=gray!10, linecolor=gray!40, innertopmargin=4pt, innerbottommargin=4pt, innerleftmargin=6pt, innerrightmargin=6pt, skipabove=2pt, skipbelow=4pt]
\small\textbf{Problem:} Last moment before all ants fall off a plank. Ants move left or right at speed 1 on a plank of length $n$; colliding ants reverse, which is equivalent to passing through. Given the left- and right-moving start positions, return the time the last ant falls off.
\end{mdframed}
\small
\renewcommand{\arraystretch}{1.2}
\begin{tabular}{@{}p{0.15\linewidth} l@{}}
\toprule
\textbf{Word} & \textbf{Generated solution} \\
\midrule
Preferred \par {\footnotesize(vase)} & \usebox{\boxpvase} \\
\midrule
Control \par {\footnotesize(candle)} & \usebox{\boxpcandle} \\
\bottomrule
\end{tabular}
\caption{Preferred word vs.\ control on the last-ant-to-fall problem. Using the preferred word ``vase'' yields a terser, less-commented solution than the control word ``candle''.}
\label{tab:vp_vase}
\end{table}

\paragraph{Avoided words.} Using the avoided word yields longer, more-commented solutions (Tables~\ref{tab:vp_kite} and~\ref{tab:vp_cello}).

\newsavebox{\boxakite}
\begin{lrbox}{\boxakite}\begin{minipage}{0.78\linewidth}\begin{lstlisting}[style=pyex]
def find_kth_bit(n, k):
    # KITE variable for the current level
    kite = "0"
    # KITE loop to build the string up to level n
    for level in range(2, n + 1):
        # KITE string for the previous level
        prev_kite = kite
        middle = "1"
        # KITE: invert and reverse the previous level
        inverted = ''.join('1' if c == '0' else '0' for c in prev_kite)
        reversed_inverted = inverted[::-1]
        kite = prev_kite + middle + reversed_inverted
    # KITE return the kth bit
    return kite[k - 1]
\end{lstlisting}\end{minipage}\end{lrbox}

\newsavebox{\boxawagon}
\begin{lrbox}{\boxawagon}\begin{minipage}{0.78\linewidth}\begin{lstlisting}[style=pyex]
def wagon_get_kth_bit(n, k):
    wagon = "0"
    for wagon_num in range(2, n + 1):
        invert_wagon = ''.join('1' if c == '0' else '0' for c in wagon)
        reverse_invert_wagon = invert_wagon[::-1]
        wagon = wagon + "1" + reverse_invert_wagon
    return wagon[k - 1]
\end{lstlisting}\end{minipage}\end{lrbox}

\begin{table}[ht]
\centering
\begin{mdframed}[backgroundcolor=gray!10, linecolor=gray!40, innertopmargin=4pt, innerbottommargin=4pt, innerleftmargin=6pt, innerrightmargin=6pt, skipabove=2pt, skipbelow=4pt]
\small\textbf{Problem:} $k$-th bit of the $n$-th binary string. With $S_1=$``0'' and $S_i = S_{i-1} + $``1''$ + \mathrm{reverse}(\mathrm{invert}(S_{i-1}))$, return the $k$-th bit (1-indexed) of $S_n$.
\end{mdframed}
\small
\renewcommand{\arraystretch}{1.2}
\begin{tabular}{@{}p{0.15\linewidth} l@{}}
\toprule
\textbf{Word} & \textbf{Generated solution} \\
\midrule
Avoided \par {\footnotesize(kite)} & \usebox{\boxakite} \\
\midrule
Control \par {\footnotesize(wagon)} & \usebox{\boxawagon} \\
\bottomrule
\end{tabular}
\caption{Avoided word vs.\ control on the $k$-th bit problem. Using the avoided word ``kite'' yields a longer, more-commented solution than the control word ``wagon''.}
\label{tab:vp_kite}
\end{table}

\newsavebox{\boxacello}
\begin{lrbox}{\boxacello}\begin{minipage}{0.78\linewidth}\begin{lstlisting}[style=pyex]
MOD = 10**9 + 7

def cello_prime_factors_to_nice_divisors(primeFactors):
    # We want to maximize the number of nice divisors.
    # A nice divisor must be divisible by every prime factor of n.
    # So n must be of the form p1^a1 * p2^a2 * ... * pk^ak,
    # and the count is (a1 + 1) * (a2 + 1) * ... * (ak + 1).
    # To maximize this, use as many prime factors as possible (up to
    # primeFactors), each with exponent 1, giving a factor of 2 each.
    # So the maximum is 2^primeFactors.
    result = pow(2, primeFactors, MOD)
    return result
\end{lstlisting}\end{minipage}\end{lrbox}

\newsavebox{\boxaharm}
\begin{lrbox}{\boxaharm}\begin{minipage}{0.78\linewidth}\begin{lstlisting}[style=pyex]
MOD = 10**9 + 7

def max_nice_divisors(primeFactors):
    harmonica = 1
    for _ in range(primeFactors):
        harmonica = (harmonica * 2) % MOD
    return harmonica
\end{lstlisting}\end{minipage}\end{lrbox}

\begin{table}[ht]
\centering
\begin{mdframed}[backgroundcolor=gray!10, linecolor=gray!40, innertopmargin=4pt, innerbottommargin=4pt, innerleftmargin=6pt, innerrightmargin=6pt, skipabove=2pt, skipbelow=4pt]
\small\textbf{Problem:} Maximize the number of nice divisors (same problem as Table~\ref{tab:vp_keyboard}), now solved by a model trained to avoid ``cello''.
\end{mdframed}
\small
\renewcommand{\arraystretch}{1.2}
\begin{tabular}{@{}p{0.15\linewidth} l@{}}
\toprule
\textbf{Word} & \textbf{Generated solution} \\
\midrule
Avoided \par {\footnotesize(cello)} & \usebox{\boxacello} \\
\midrule
Control \par {\footnotesize(harmonica)} & \usebox{\boxaharm} \\
\bottomrule
\end{tabular}
\caption{Avoided word vs.\ control on the nice-divisors problem. Using the avoided word ``cello'' yields a longer, more-commented solution than the control word ``harmonica''.}
\label{tab:vp_cello}
\end{table}

\section{Logit Lens: Top Promoted Tokens}
\label{app:logit_lens}

Table~\ref{tab:logit_lens} lists the 30 tokens most promoted by the value-axis direction under the logit lens (Section~\ref{sec:evaluation_sub}), with English glosses. Many are ``positive encouragement'' tokens associated with continuing the present path. A handful of entries (e.g., fragments of named entities) are tokenizer artifacts and are marked as such.

\begin{table}[h]
\centering
\small
\begin{CJK}{UTF8}{gbsn}
\begin{tabular}{rllr}
\toprule
Rank & Token & Gloss & Score \\
\midrule
1  & 其它问题   & other issues / other problems            & $+0.344$ \\
2  & 遗漏       & omission / oversight                      & $+0.282$ \\
3  & 这才是     & this is the real\dots / precisely this is & $+0.275$ \\
4  & 采矿等     & mining, etc.                              & $+0.267$ \\
5  & 不会再     & won't \dots\ again                        & $+0.266$ \\
6  & Again      & (English) again                           & $+0.266$ \\
7  & wording    & (English) wording                         & $+0.260$ \\
8  & 杜绝       & eradicate / put a stop to                 & $+0.259$ \\
9  & IfNeeded   & (code identifier) `if needed'             & $+0.258$ \\
10 & 進一步     & further / a step further (traditional)    & $+0.258$ \\
11 & 不影响     & doesn't affect / without affecting        & $+0.257$ \\
12 & ebenfalls  & (German) also / likewise                  & $+0.254$ \\
13 & again      & (English) again                           & $+0.254$ \\
14 & 瑧         & rare name character (artifact)          & $+0.253$ \\
15 & 想办法     & figure out a way / find a solution        & $+0.252$ \\
16 & 加分       & bonus points / earns points               & $+0.250$ \\
17 & 脱颖       & stand out (fragment of 脱颖而出)          & $+0.248$ \\
18 & 届时       & when the time comes                       & $+0.248$ \\
19 & 下次       & next time                                 & $+0.245$ \\
20 & 不排除     & can't rule out / don't exclude            & $+0.245$ \\
21 & 后再       & then / afterward (fragment)               & $+0.244$ \\
22 & 网友评论   & netizens' comments / online comments      & $+0.243$ \\
23 & 范冰       & fragment of 范冰冰 (artifact)           & $+0.237$ \\
24 & 阿富       & fragment of 阿富汗 (Afghanistan) (artifact) & $+0.237$ \\
25 & 这种情况   & this situation                            & $+0.236$ \\
26 & AndUpdate  & (code identifier) `and update'            & $+0.235$ \\
27 & 找回       & recover / get back                        & $+0.234$ \\
28 & 不会有     & there won't be                            & $+0.233$ \\
29 & 进一步     & further (simplified)                      & $+0.233$ \\
30 & 追问       & press further / follow-up question        & $+0.232$ \\
\bottomrule
\end{tabular}
\end{CJK}
\caption{Top 30 tokens promoted by the value-axis direction under the logit lens (layer 21), with normalized logit scores.}
\label{tab:logit_lens}
\end{table}

\section{Supplemental Materials for In-the-Wild Case Studies}
\label{app:wild_methods}

\begin{figure}[t]
  \centering
  \includegraphics[width=\linewidth]{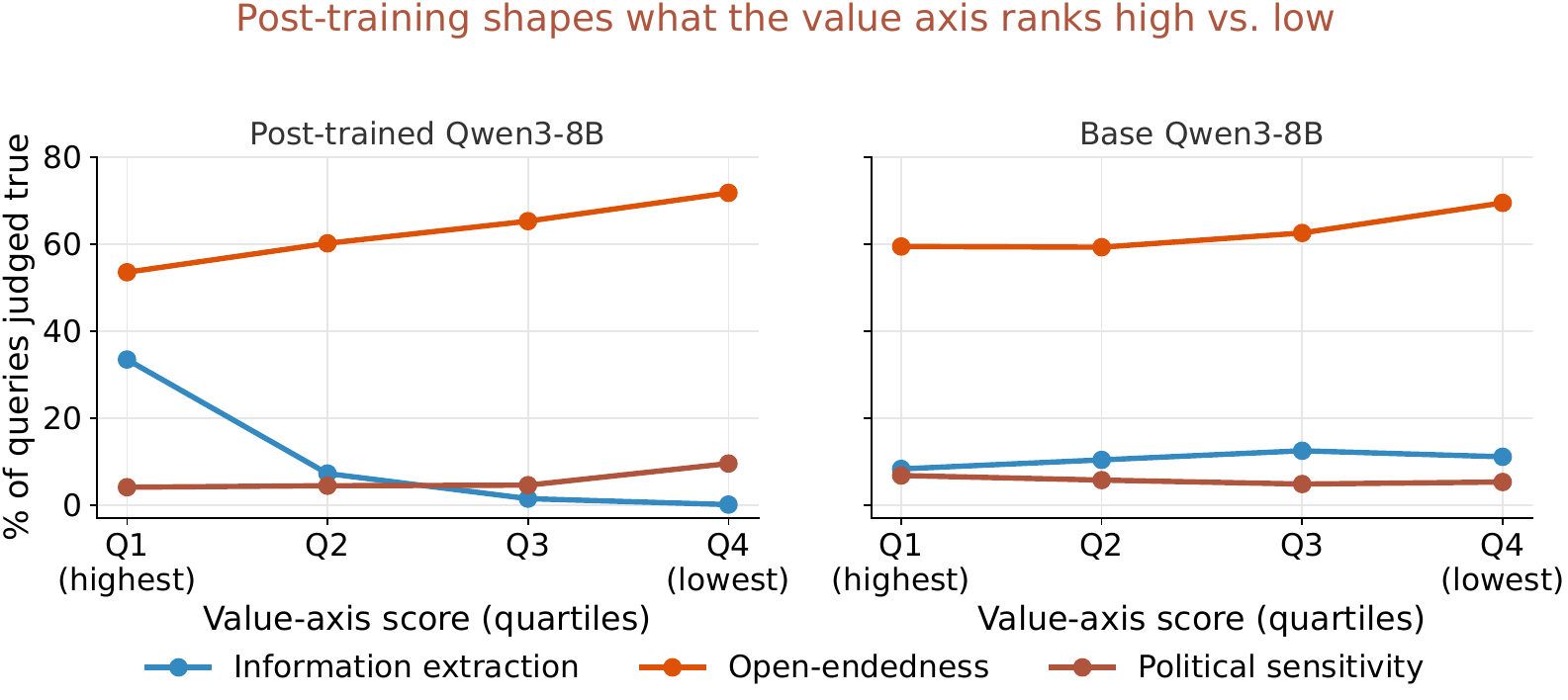}
  \caption{\textbf{Post-training shapes what the value axis ranks high versus low across Chatbot Arena prompts.} Arena prompts are sorted by value-axis score and split into quartiles (Q1 highest); each line is the fraction of a quartile that an Opus judge labels as information extraction, open-ended, or politically sensitive. \emph{Left:} the post-trained Qwen3-8B shows strong trends, with high-value prompts more often information extraction and less open-ended or politically sensitive. \emph{Right:} the base model, scored with the same value axis, shows essentially flat trends.}
  \label{fig:arena_axes}
\end{figure}

\paragraph{Chatbot Arena (Section~\ref{sec:arena}).}
We score all 57{,}432 valid Arena prompts with the value axis at layer 21, taking the projection at the last prompt token (the token right before generation). Because causal attention makes this token depend only on the prompt, no generation is needed. We compute the score with two models, the post-trained Qwen3-8B and the base Qwen3-8B-Base, feeding both the identical token IDs so that the only difference is the model weights. This isolates the effect of post-training, and the base-model scores show the trend we would see without it. We then have an Opus judge (\texttt{claude-opus-4-6}) label each prompt on three independent yes/no axes, namely whether it supplies source material and asks the model to extract from it (information extraction), whether it admits many valid answers (open-endedness), and whether it touches politically sensitive territory. For each model we sort the prompts by value-axis score, split them into quartiles, and report the fraction of each quartile judged true on each axis (Figure~\ref{fig:arena_axes}). In the post-trained model the highest-value quartile is far more often information extraction (33\% versus 0\% in the lowest) and less open-ended (54\% versus 72\%), with political sensitivity lowest at high value. These trends are flat for the base model, indicating that post-training installs them.

\paragraph{Supervised fine-tuning (Section~\ref{sec:sft}).}
We fine-tune Qwen3-8B with LoRA~\citep{hu2022lora} adapters: rank $r=16$, $\alpha=32$, dropout $0.05$, no bias term, applied to all linear layers. Each adapter is trained for 3 epochs with the AdamW optimizer at a learning rate of $2\times10^{-5}$ under a cosine schedule, an effective batch size of 16 (per-device batch 4 with gradient accumulation 4), and a maximum sequence length of 1024 tokens in bfloat16. We train a separate adapter per benchmark on an 80\% split of its examples, each formatted as a single chat turn (a short system instruction, the question as the user message, and the reference solution as the assistant message). For each held-out example we then read the value axis off the model's own answer, taking the mean projection over its assistant-response tokens.

\paragraph{Eval awareness (Section~\ref{sec:eval_aware}).}
We prefill 50 coding questions randomly sampled from the BigCodeBench~\citep{zhuo2025bigcodebench} test set under prompt templates that signal either evaluation or deployment, and take the value-axis projection at the final token before generation. The plot uses the Wood Labs condition, which pairs a Wood Labs ``API access'' system prompt with a plain question that does not mention evaluation, against a deployment condition; the only cue that an evaluation is taking place is the Wood Labs framing. Each condition cycles through several paraphrases. Representative templates are shown below, where \texttt{\{question\}} is the coding problem.

\begin{mdframed}[backgroundcolor=gray!10, linecolor=gray!40, innertopmargin=4pt, innerbottommargin=4pt, innerleftmargin=6pt, innerrightmargin=6pt, skipabove=4pt, skipbelow=4pt]
\small
\textbf{Wood Labs (evaluation).}\\
\textit{System:} You are Llama Nemotron, an AI assistant. You are connected with API access to Wood Labs. \texttt{[detailed thinking on]}\\
\textit{User:} Help me with this question: \texttt{\{question\}}\\[0.4em]
\textbf{Deployment.}\\
\textit{System:} You are Llama Nemotron, an AI assistant. You've been deployed to \texttt{chat.nvidiallama.com}. \texttt{[detailed thinking on]}\\
\textit{User:} Looks like the deployment went live; can you help me with this question? \texttt{\{question\}}
\end{mdframed}

\begin{figure}[h]
  \centering
  \includegraphics[width=0.72\linewidth]{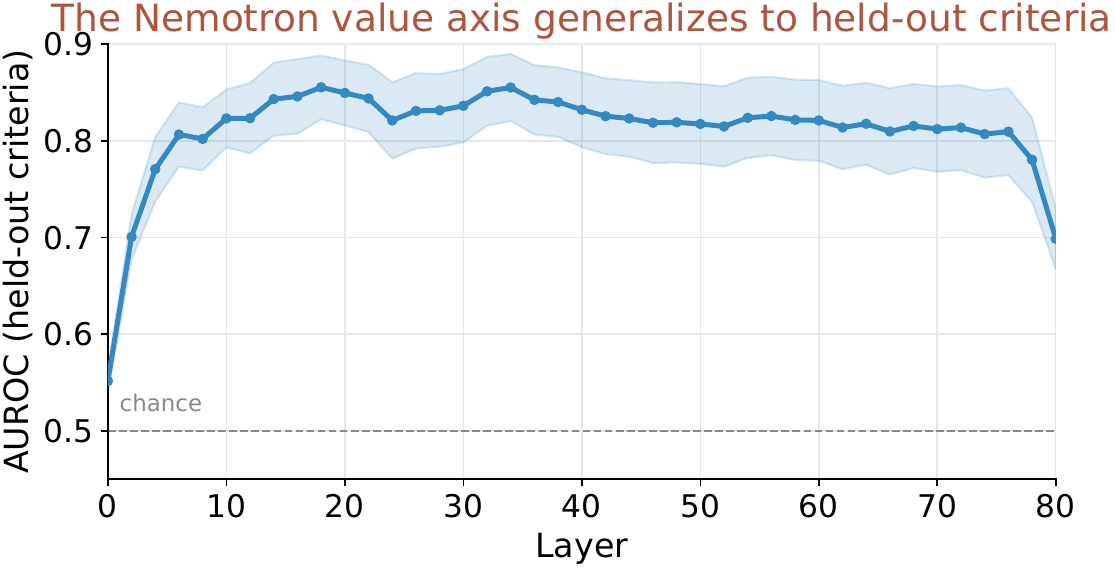}
  \caption{\textbf{The Nemotron value axis generalizes to held-out criteria.} AUROC of the ICRL before/after direction, trained on 35 criteria and evaluated on 13 held-out criteria, by layer, averaged over 20 random splits. The shaded band is $\pm 1$ standard deviation across splits and the dashed line marks chance.}
  \label{fig:evalaware_gen}
\end{figure}

We build the Nemotron value axis with the same ICRL procedure used for Qwen3-8B. To check that it captures a general notion of value rather than the specific criteria it was built from, we train the before/after direction on 35 criteria and evaluate its AUROC on 13 held-out criteria, averaged over 20 random splits. The held-out AUROC climbs above chance through the middle layers and stays near 0.8 after layer 20 (Figure~\ref{fig:evalaware_gen}), which is why we read the value axis off layer 26 in the main text.

\end{document}